\documentclass{article}
\pdfoutput=1
\usepackage[nonatbib,final]{neurips_2021}
\usepackage[utf8]{inputenc} % allow utf-8 input
\usepackage[T1]{fontenc}    % use 8-bit T1 fonts
\usepackage{color,xcolor}
\usepackage{epsfig}
\usepackage{graphicx}
\usepackage{grffile} 

\usepackage{adjustbox}
\usepackage{array}
\usepackage{booktabs}
\usepackage{colortbl}
\usepackage{float,wrapfig}
\usepackage{hhline}
\usepackage{multirow}
\usepackage{subcaption} % issues a warning with CVPR/ICCV format
\usepackage{caption}
\usepackage{amsmath,amsfonts,amsthm,amssymb}
\usepackage{bm}
\usepackage{nicefrac}
\usepackage{microtype}

\usepackage{changepage}
\usepackage{extramarks}
\usepackage{fancyhdr}
\usepackage{lastpage}
\usepackage{setspace}
\usepackage{soul}
\usepackage{xspace}

\usepackage[pagebackref=true,breaklinks=true,colorlinks,citecolor=gray]{hyperref}
\usepackage[nocompress]{cite}
\usepackage{url}

\usepackage{algorithm, algorithmic}
\usepackage{enumerate}
\usepackage{todonotes} % conflict with CVPR/ICCV format
\usepackage[frozencache,cachedir=minted-cache]{minted}
\setminted[python]{
linenos,
numbersep=5pt,
frame=lines,
bgcolor=lightgray!10,
framesep=2mm,
baselinestretch=1.2,
fontsize=\footnotesize,
framerule=0.8pt
}

\usepackage{titlesec}
\usepackage{listings}
\usepackage{enumitem}
\usepackage{numprint}

\usepackage{pifont}% http://ctan.org/pkg/pifont
\usepackage{tabularx}
\newcolumntype{L}{>{\arraybackslash}X}
\newcolumntype{C}{>{\centering\arraybackslash}X}

\newcommand{\revise}[1]{\textcolor{black}{#1}}
\newcommand\redsout{\bgroup\markoverwith{\textcolor{red}{\rule[0.5ex]{2pt}{0.4pt}}}\ULon}

\newcommand{\libname}{RLlib\xspace}
\newcommand{\flowname}{RLlib Flow\xspace}
\newcommand{\flownamecap}{RLlib Flow\xspace}

\newcommand{\tab}[1]{Table~\ref{tab:#1}}
\newcommand{\fig}[1]{Figure~\ref{fig:#1}}
\newcommand{\code}[1]{Figure~\ref{alg:#1}}
\newcommand{\lst}[1]{Listing~\ref{lst:#1}}
\newcommand{\sect}[1]{Section~\ref{sec:#1}}

\newcommand{\apdx}[1]{Appendix~\ref{append:#1}}
\newcommand{\x}{$\times$}
\newenvironment{mycode}{\captionsetup{type=listing}}{}

\title{\flownamecap: Distributed Reinforcement Learning is a Dataflow Problem}

\author{%
  Eric Liang$^{*}$ \\
  UC Berkeley
   \And
   Zhanghao Wu$^{*}$ \\
   UC Berkeley
   \And
   Michael Luo \\
   UC Berkeley
   \AND
   Sven Mika \\
   Anyscale
   \And
   Joseph E. Gonzalez \\
   UC Berkeley
    \And
   Ion Stoica \\
   UC Berkeley
}

\begin{document}
\maketitle

\footnotetext{$*$ indicates equal contributions.}

\begin{abstract}
Researchers and practitioners in the field of reinforcement learning (RL) frequently leverage parallel computation, which has led to a plethora of new algorithms and systems in the last few years. In this paper, we re-examine the challenges posed by distributed RL and try to view it through the lens of an old idea: distributed dataflow. We show that viewing RL as a dataflow problem leads to highly composable and performant implementations.
We propose \flowname, a hybrid actor-dataflow programming model for distributed RL, and validate its practicality by porting the full suite of algorithms in \libname, a widely adopted distributed RL library. \revise{Concretely, \flowname provides 2-9$\times$ code savings in real production code and enables the composition of multi-agent algorithms not possible by end users before. The open-source code is available as part of RLlib at \url{ https://github.com/ray-project/ray/tree/master/rllib}.}
\end{abstract}

\section{Introduction}
\label{introduction}

The past few years have seen the rise of deep reinforcement learning (RL) as a new, powerful optimization method for solving sequential decision making problems. As with deep supervised learning, researchers and practitioners frequently leverage parallel computation, which has led to the development of numerous distributed RL algorithms and systems as the field rapidly evolves.

However, despite the high-level of abstraction that RL algorithms are defined in (i.e., as a couple dozen lines of update equations), their implementations have remained quite low level (i.e., at the level of message passing). This is particularly true for \textit{distributed} RL algorithms, which are typically implemented directly on low-level message passing systems or actor frameworks \cite{hewitt1973session}. Libraries such as Acme \cite{hoffman2020acme}, RLgraph \cite{Schaarschmidt2019rlgraph}, RLlib \cite{liang2018rllib}, and Coach \cite{coach} provide unified abstractions for defining single-agent RL algorithms, but their user-facing APIs only allow algorithms to execute within the bounds to predefined distributed execution patterns or ``templates''.

While the aforementioned libraries have been highly successful at replicating a large number of novel RL algorithms introduced over the years, showing the generality of their underlying actor or graph-based computation models, the needs of many researchers and practitioners are often not met by their abstractions. We have observed this firsthand from users of open source RL libraries:

First, RL practitioners are typically not systems engineers. They are not well versed with code that mixes together the logical dataflow of the program and system concerns such as performance and bounding memory usage. This leads to a high barrier of entry for most RL users to experimenting with debugging existing distributed RL algorithms or authoring new distributed RL approaches.

Second, even when an RL practitioner is happy with a particular algorithm, they may wish to \textit{customize} it in various ways. This is especially important given the diversity of RL tasks (e.g., single-agent, multi-agent, meta-learning). While many customizations within common RL environments can be anticipated and made available as configuration options (e.g., degree of parallelism, batch size), it is difficult for a library author to provide enough options to cover less common tasks that necessarily alter the distributed pattern of the algorithm (e.g., interleaved training of different distributed algorithms, different replay strategies).

%\michael{Third, implementing distributed execution from scratch is complicated and messy!It's good to have a modular framework...}

% https://docs.google.com/drawings/d/11pJ7KkblxeipjecAqdBOk-iZUZid43dItsFt3Bd8TYA/edit
\begin{figure}[t]
\centering
\begin{minipage}[b]{0.45\linewidth}
\includegraphics[width=.98\linewidth]{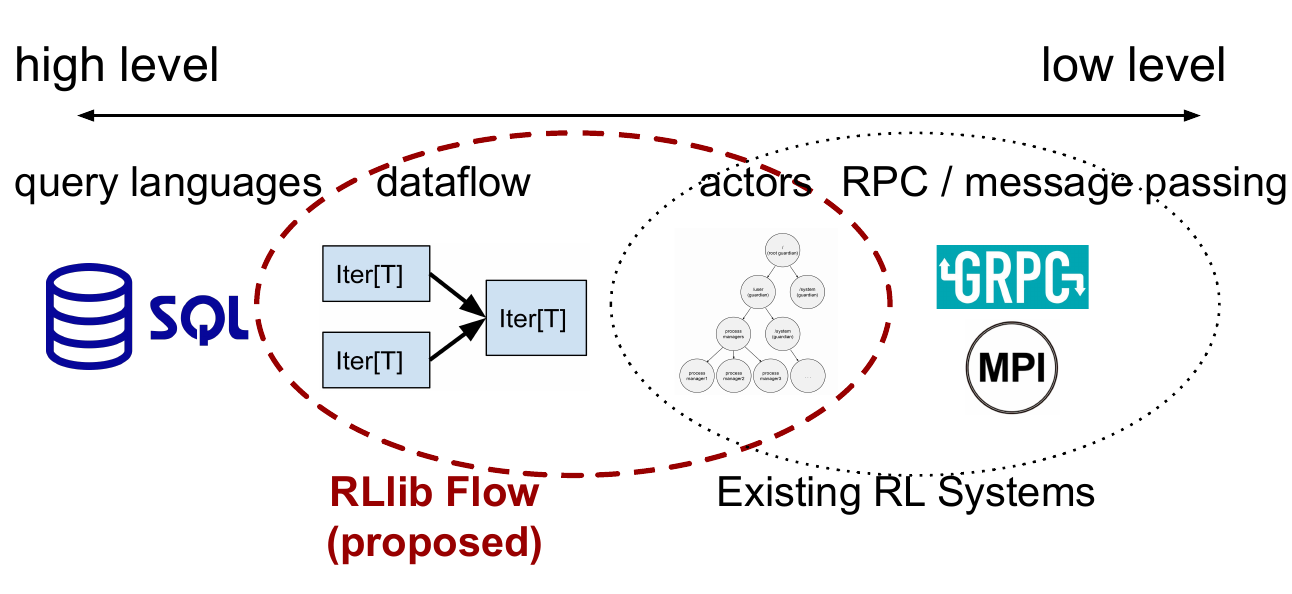}
\vspace{-10pt}
\caption{\flowname enables implementation of distributed RL with high-level dataflow.}
\label{fig:dataflow}
\end{minipage}
~
\begin{minipage}[b]{0.52\linewidth}
% https://docs.google.com/document/d/1mlPrUrlPxiT7rtQFLDhfUuJwo6L9IZbyCI_SM6_qTqE/edit#
\includegraphics[width=.98\linewidth]{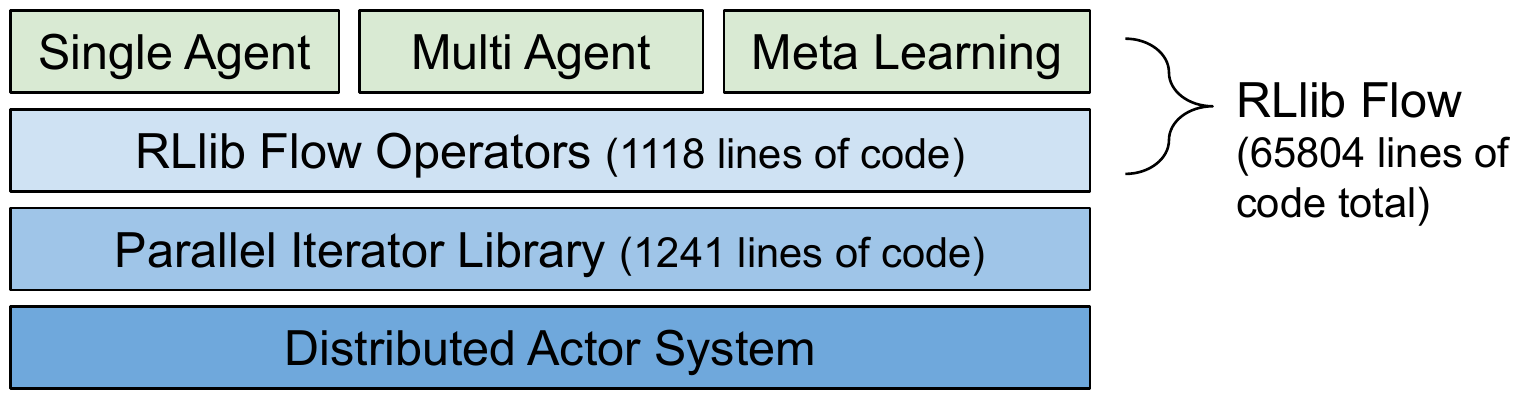}
\vspace{2pt}
\caption{Architecture of our port of \libname to \flowname (\flowname core is only 1118 lines of code).}
\label{fig:arch}
\end{minipage}
% \vspace{-10pt}
% \caption{We propose \flowname, a hybrid actor-dataflow model fordistributed RL.}
% \vspace{-15pt}
\end{figure}
\begin{table}[t]
\renewcommand*{\arraystretch}{1.3}
\setlength{\tabcolsep}{1.8pt}
\centering
\caption{\revise{Comparison of the systems aspects of \flowname to other distributed RL libraries.}}
\label{tab:capability}
{\small
\begin{tabularx}{\linewidth}{>{\raggedright}p{80pt}>{\raggedright}p{80pt}>{\raggedright}p{80pt}>{\raggedright}p{100pt}C}\toprule
Distributed Library &Distribution Scheme &Generality &Programmability &\#Algorithms \\\midrule
RLGraph~\cite{Schaarschmidt2019rlgraph} &Pluggable &General Purpose &Low-level / Pluggable &10+ \\
Deepmind Acme~\cite{hoffman2020acme} & Actors + Reverb &Async Actor-Learner &Limited &10+ \\
Intel Coach~\cite{coach} &Actors + NFS & Async Actor-Learner &Limited &30+ \\
RLlib~\cite{liang2018rllib} &Ray Actors &General Purpose &Flexible, but Low-level &20+ \\
\textbf{\flowname} &\textbf{Actor / Dataflow} &\textbf{General Purpose} &\textbf{Flexible and High-level} &\textbf{20+} \\
\bottomrule
\end{tabularx}
}
% \tabspace
% \vspace{-10pt}
\end{table}

Our experience is that when considering the needs of users for novel RL applications and approaches, RL development requires a significant degree of programming flexibility. Advanced users want to tweak or add various distributed components (i.e., they need to write programs). In contrast to supervised learning, it is more difficult to provide a fixed set of abstractions for scaling RL training.

As a result, it is very common for RL researchers or practitioners to eschew existing infrastructure, either sticking to non-parallel approaches, which are inherently easier to understand and customize \cite{dopamine, baselines}, or writing their own distributed framework that fits their needs. The large number of RL frameworks in existence today is evidence of this, especially considering the number of these frameworks aiming to be ``simpler'' versions of other frameworks.

In this paper, we re-examine the challenges posed by distributed RL in the light of these user requirements, drawing inspiration from prior work in the field of data processing and distributed dataflow.
To meet these challenges, we propose \flowname, a hybrid actor-dataflow model for distributed RL. Like streaming data systems, \flowname provides a small set of operator-like primitives that can be composed to express distributed RL algorithms. Unlike data processing systems, \flowname explicitly exposes references to actor processes participating in the dataflow, permitting limited message passing between them in order to more simply meet the requirements of RL algorithms. The interaction of dataflow and actor messages is managed via special sequencing and concurrency operators.

The contributions of our paper are as follows:
\begin{enumerate}[leftmargin=*]
\item We examine the needs of distributed RL algorithms and RL practitioners from a dataflow perspective, identifying key challenges (Section \ref{sec:challenges} and \ref{sec:streaming}).
\item We propose \flowname, a hybrid actor-dataflow programming model that can simply and efficiently express distributed RL algorithms\revise{, and enables composition of multi-agent algorithms not possible by end users before without writing low-level systems code.} (Section \ref{sec:dataflow} and \ref{sec:implementation}).
\item We port all the algorithms of a production RL library (\libname) to \flowname, providing 2-9$\times$ savings in distributed execution code, compare its performance with the original implementation, and show performance benefits over systems such as Spark Streaming (Section \ref{sec:evaluation}).
\end{enumerate}

\section{Distributed Reinforcement Learning}
\label{sec:challenges}

We first discuss the relevant computational characteristics of distributed RL algorithms, starting with the common \textit{single-agent training} scenario, where the goal is to optimize a single agent's performance in an environment, and then discuss the computational needs of emerging \textit{multi-agent}, \textit{model-based}, and \textit{meta-learning} training patterns.

\subsection{RL Algorithm Basics}

The goal of an RL algorithm is typically to improve the performance of a \textit{policy} with respect to an objective defined through an \textit{environment} (e.g., simulator). The policy is usually defined as a deep neural network, which can range from several KB to several hundred MB in size. RL algorithms can be generally broken down into the following basic steps:

%https://docs.google.com/document/d/1mlPrUrlPxiT7rtQFLDhfUuJwo6L9IZbyCI_SM6_qTqE/edit#
\begin{figure}[h]
\centering
\includegraphics[width=.5\linewidth]{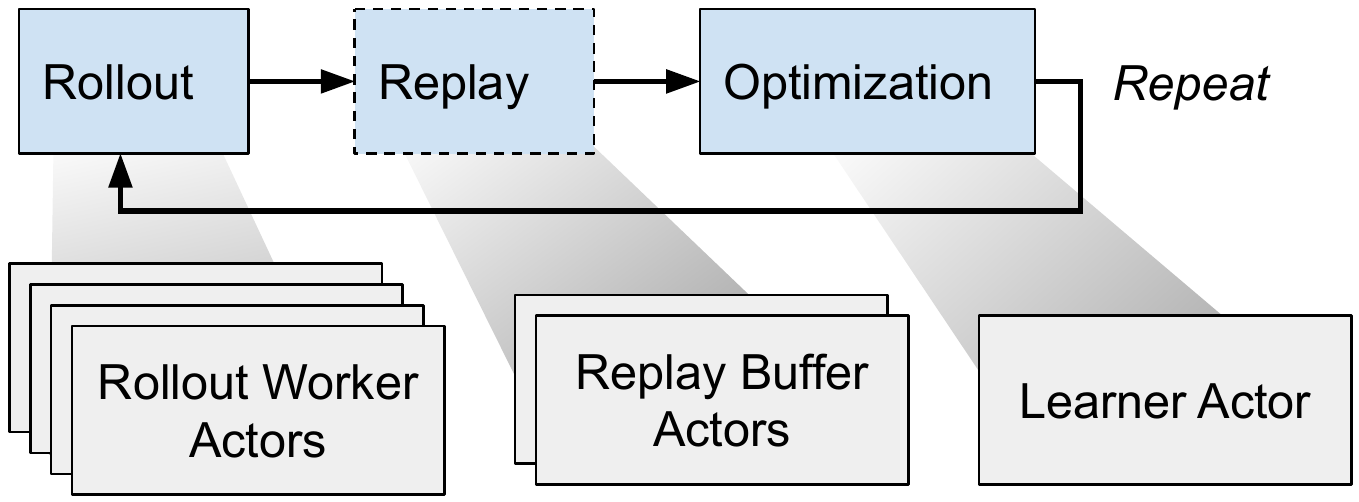}
% \vspace{-3pt}
\caption{Most RL algorithms can be defined in terms of the basic steps of Rollout, Replay, and Optimization. These steps are commonly parallelized across multiple \textit{actor} processes.}
\label{fig:rl_basic}
% \vspace{-5pt}
\end{figure}

\textbf{Rollout}: To generate experiences, the policy, which outputs actions to take given environment observations, is run against the environment to collect batches of data. The batch consists of observations, actions, rewards, and episode terminals and can vary in size (10s to 10000s of steps).% In general, rollouts may involve environments hosted on remote machines, interacting directly with the physical world, or data indirectly gathered through log files. While such scenarios require additional engineering, we consider handling them an orthogonal concern to \flowname.

\textbf{Replay}: On-policy algorithms (e.g., PPO \cite{schulman2017proximal}, A3C \cite{mnih2016asynchronous}) collect new experiences from the current policy to learn. On the other hand, off-policy algorithms (e.g., DQN \cite{mnih2015humanlevel}, SAC \cite{softlearning}) can leverage experiences from past versions of the policy as well. For these algorithms, a \textit{replay buffer} of past experiences can be used. The size of these buffers ranges from a few hundred to millions of steps.

\textbf{Optimization}: Experiences, either freshly collected or replayed, can be used to improve the policy. Typically this is done by computing and applying a gradient update to the policy and value neural networks. While in many applications a single GPU suffices to compute gradient updates, it is sometimes desirable to leverage multiple GPUs within a single node, asynchronous computation of gradients on multiple CPUs \cite{mnih2016asynchronous}, or many GPUs spread across a cluster \cite{wijmans2020ddppo}.

\subsection{RL Algorithm Variants}
\textbf{Single-Agent Training.}
Training a single RL agent---the most basic and common scenario---consists of applying the steps of rollout, replay, and optimization repeatedly until the policy reaches the desired performance. Synchronous algorithms such as A2C \cite{mnih2016asynchronous} and PPO apply the steps strictly sequentially. Parallelism may be leveraged internally within each step. Asynchronous algorithm variations such as A3C \cite{mnih2016asynchronous}, Ape-X \cite{horgan2018distributed
}, APPO \cite{luo2020impact}, and IMPALA \cite{espeholt2018impala}, pipeline and overlap the rollout and optimization steps asynchronously to hit higher data throughputs. Rate limiting~\cite{hoffman2020acme} can be applied to control learning dynamics in the asynchronous setting.

\textbf{Multi-Agent Training.}
In multi-agent training, there are multiple acting entities in the environment (e.g., cooperating or competing agents). While there is a rich literature on multi-agent algorithms, we note that the \textit{dataflow structure} of multi-agent training is similar to that of single-agent---as long as all entities are being trained with the same algorithm and compatible hyperparameters.% Intuitively, for a single algorithm, the changes necessary to support multi-agent consist of replicating the rollout, replay, and optimization steps per agent.
However, problems arise should it be required to customize the training of any of the agents in the environment. For example, in a two-agent environment, one agent may desire to be optimized at a higher frequency (i.e., smaller batch size). This fundamentally alters the training dataflow---there are now two iterative loops executing at different frequencies. Furthermore, if these agents are trained with entirely different algorithms, there is a need to compose two different distributed dataflows.% We have seen several users of \libname encounter both of these scenarios in their work. % They are quite difficult to address by adding configuration options since it is difficult to anticipate how users want to customize their training workflow.

\textbf{Model-Based and Meta-Learning Algorithms.}
Model-based algorithms seek to learn transition dynamics of the environment to improve the sample efficiency of training. This can be thought of as adding a supervised training step on top of standard distributed RL, where an ensemble of one or more dynamics models are trained from environment-generated data. Handling the data routing, replay, optimization, and stats collection for these models naturally adds complexity to the distributed dataflow graph, ``breaking the mold'' of standard model-free RL algorithms and hard to be implemented in low-level systems. Using \flowname, we have implemented two state-of-the-art model-based algorithms: MB-MPO \cite{mbmpo} and Dreamer \cite{hafner2020dream}.% Similarly, meta-learning algorithms such as MAML \cite{maml} leverage additional computation structures (i.e., nested inner optimization steps) as they seek to learn a policy amenable for quick adaptation to new environments.% Through \flowname we are able to support both  MAML \cite{maml} in MB-MPO.% The proposed dataflow DSL enabled the optimization step to be easily customized to support the optimization of an ensemble of models distributed across many different worker processes.

\subsection{A Case for a Higher Level Programming Model}

Given that existing distributed RL algorithms are already implementable using low level actor and RPC primitives, it is worth questioning the value of defining a higher level computation model. Our experience is that RL is more like data analytics than supervised learning. Advanced users want to tweak or add various distributed components (i.e., they need to program), and there is no way to have a ``one size fits all'' (i.e., Estimator interface from supervised learning). We believe that, beyond the ability to more concisely and cleanly capture \textit{single-agent} RL algorithms, the computational needs of more advanced RL training patterns motivate higher level programming models like \flowname.
\section{Reinforcement Learning vs Data Streaming}
\label{sec:streaming}

The key observation behind \flowname is that the dataflow graph of RL algorithms are quite similar to those of data streaming applications. Indeed, RL algorithms can be captured in general purpose dataflow programming models. However, 
%as we will see, 
due to several characteristics, they are not a perfect fit, even for dataflow programming models that support iterative computation.

%Both distributed RL and data streaming share several commonalities such as a regularly recurring computational structure, the need for high performance, and a certain level of programmability to handle novel use cases. However, they differ in requirements around fault tolerance, consistency, correctness, and durability. Reinforcement learning also heavily leverages ``iterative'' computation, which is less common in data-oriented systems.

%\subsection{Example: Asynchronous Actor Critic (A3C)}

\begin{figure}[h]
 %https://docs.google.com/presentation/d/10bLdyeErPKxxyCNrzhGdHG91L6F82Y_OnomxN4eIwUY/edit#slide=id.ga037b6c774_0_0\begin{figure}
\centering
\includegraphics[width=0.8\linewidth]{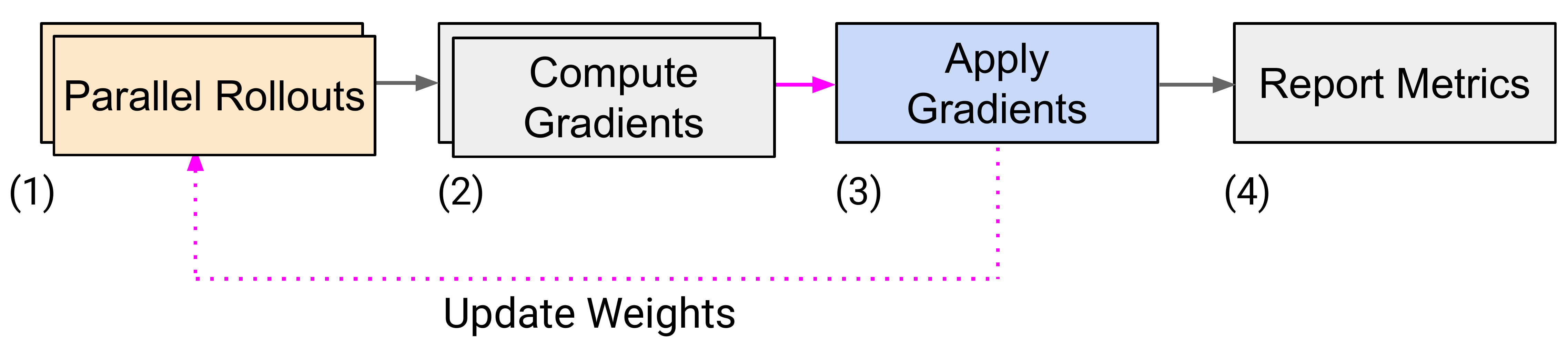}
\caption{The dataflow of the A3C parallel algorithm. Each box is an operator or iterator from which data items can be pulled from. Here operators (1) and (2) represent parallel computations, but (3) and (4) are sequential. Black arrows denote synchronous data dependencies, pink arrows asynchronous dependencies, and dotted arrows actor method calls.% Training metrics are pulled from the output operator ("Report Metrics"), which drives the computation.
}
\label{fig:a3c}
% \vspace{-10pt}
\end{figure}

In this section we examine the dataflow of the A3C algorithm (Figure \ref{fig:a3c}) to compare and contrast RL with streaming dataflow. A3C starts with (1) parallel rollouts across many experiences. Policy gradients are computed in parallel based on rollouts in step (2). In step (3), the gradients are asynchronously gathered and applied on a central model, which is then used to update rollout worker weights. Importantly, each box or \textit{operator} in this dataflow may be \textit{stateful} (e.g., \texttt{ParallelRollouts} holds environment state as well as the current policy snapshot).

%Taken altogether, the sub-flow defined by (1)-(2) execute independently in parallel. Their gradient outputs are gathered asynchronously (solid pink arrow) and then applied to update the policy in (3), which also updates the policy weights of the actors generating the rollouts in (1) (dotted pink arrow). We compare the A3C dataflow graph with that of data streaming jobs:

%\textbf{Computation Structure}:
Similar to data processing topologies, A3C is applying a transformation to a data stream (of rollouts) in parallel (to compute gradients). This is denoted by the black arrow between (1) and (2). There is also a non-parallel transformation to produce metrics from the computation, denoted by the black arrow between (3) and (4) However, zooming out to look at the entire dataflow graph, a few differences emerge:

\textbf{Asynchronous Dependencies}: RL algorithms often leverage asynchronous computation to reduce update latencies and eliminate stragglers \cite{mnih2016asynchronous}. In \flowname, we represent these with a pink arrow between a parallel and sequential iterator. This means items will be fetched into the sequential iterator as soon as they are available, instead of in a deterministic ordering. The level of asynchrony can be configured to increase pipeline parallelism. 

\textbf{Message Passing}: RL algorithms, like all iterative algorithms, need to update upstream operator state during execution (e.g., update policy weights). Unlike iterative algorithms, these updates may be fine-grained and asynchronous (i.e., update the parameters of a particular worker), as well as coarse-grained (i.e., update all workers at once after a global barrier). \flowname allows method calls (messages) to be sent to any actor in the dataflow. Ordering of messages in \flowname with respect to dataflow steps is guaranteed if synchronous data dependencies (black arrows) fully connect the sender to the receiver, providing \textit{barrier semantics}.

%\textbf{Performance}: Similar to data processing, RL is data and compute intensive. This motivates a programming model that minimizes overhead (e.g., state serialization and logging cost).

\textbf{Consistency and Durability}: Unlike data streaming, which has strict requirements such as exactly-once processing of data~\cite{zaharia_spark_stream_2013}, RL has less strict consistency and durability requirements. This is since on a fault, the entire computation can be restarted from the last checkpoint with minimal loss of work. Message or data loss can generally be tolerated without adverse affect on training. Individual operators can be restarted on failure, discarding any temporary state. This motivates a programming model that minimizes overhead (e.g., avoids state serialization and logging cost).

%\textbf{Consistency}: Unlike data streaming, which has strict requirements such as exactly-once processing of data, in RL data can generally be discarded without much adverse affect, as long as the data isn't too stale as to destabilize learning.

\section{A Dataflow Programming Model for Distributed RL}
\label{sec:dataflow}

Here we formally define the \flowname hybrid actor-dataflow programming model. \flowname consists of a set of dataflow operators that produce and consume \textit{distributed iterators} \cite{volcano}. These distributed iterators can represent parallel streams of a data items \texttt{T} sharded across many actors (\texttt{ParIter[T]}), or a single sequential stream of  items (\texttt{Iter[T]}).
It is important to note that these iterators are \textit{lazy}, they do not execute computation or produce items unless requested. This means that the entire \flowname execution graph driven by taking items from the output operator.

\begin{figure}[h]
\centering
\begin{minipage}[b]{.48\linewidth}
{\scriptsize
\begin{lstlisting}
create(Seq[SourceActor[T]]) $\rightarrow$ ParIter[T]
send_msg(dest: Actor, msg: Any) $\rightarrow$ Reply
\end{lstlisting}
}
\vspace{-4pt}
\centering
\includegraphics[width=0.8\linewidth]{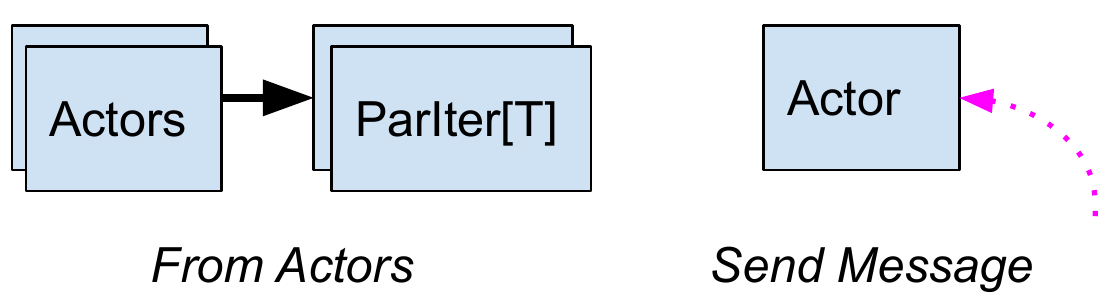}
\caption{Creation and Message Passing}
\label{fig:creation}
\end{minipage}
~
\begin{minipage}[b]{.48\linewidth}
{\scriptsize
\begin{lstlisting}
for_each(ParIter[T], T => U) $\rightarrow$ ParIter[U]
for_each(Iter[T], T => U) $\rightarrow$ Iter[U]
\end{lstlisting}
}
\centering
\includegraphics[width=0.92\linewidth]{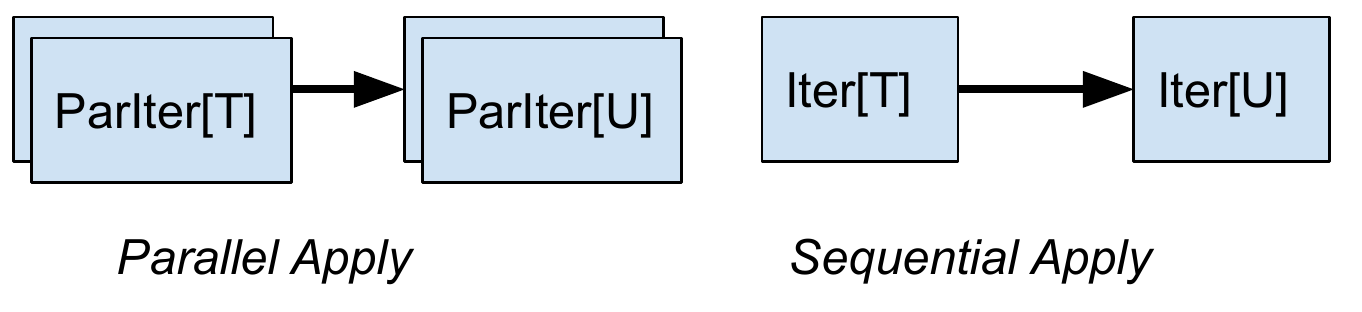}
\vspace{-4pt}
\caption{Transformation}
\label{fig:transformation}
\end{minipage}
\end{figure}

\textbf{Creation and Message Passing}: \flowname iterators are always created from an existing set of actor processes. In \fig{a3c}, the iterator is created from a set of rollout workers that produce experience batches given their current policy. Also, any operator may send a message to any source actor (i.e., a rollout worker, or replay buffer) during its execution. In the A3C example, the update weights operation is a use of this facility. The order guarantees of these messages with respect to dataflow steps depends on the barrier semantics provided by \textit{sequencing operators}. The sender may optionally block and await the reply of sent messages. We show the operator in~\fig{creation}.

\textbf{Transformation}: As in any data processing system, the basic operation of data transformation is supported. Both parallel and sequential iterators can be transformed with the \texttt{for\_each} operator. The transformation function can be stateful (i.e., in Python it can be a callable function class that holds state in class members, and in the case of sequential operators it can reference local variables via closure capture). In the A3C example, \texttt{for\_each} is used to compute gradients for each batch of experiences, which depends on the current policy state of the source actor. In the case of the \texttt{ComputeGradients} step, this state is available in the local process memory of the rollout worker, and is accessible because \flowname schedules the execution of parallel operations onto the source actors. We show the operator in~\fig{transformation}.

\begin{figure}[h]
\begin{minipage}{.49\linewidth}
{\scriptsize
\begin{lstlisting}
gather_async(ParIter[T], 
             num_async: Int) $\rightarrow$ Iter[T]
gather_sync(ParIter[T]) $\rightarrow$ Iter[List[T]]
next(Iter[T]) $\rightarrow$ T
\end{lstlisting}
}
\centering
\includegraphics[width=0.8\linewidth]{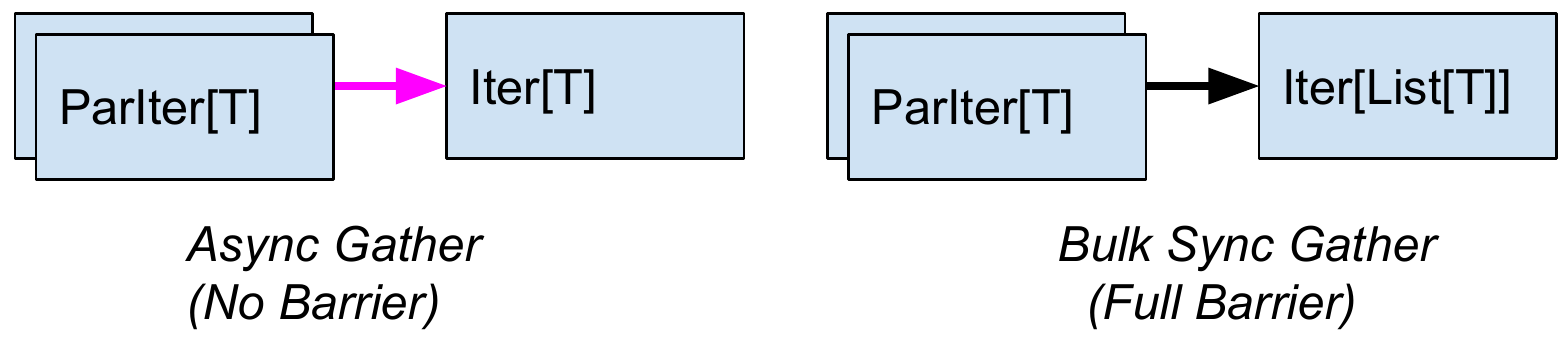}
\caption{Sequencing}
\label{fig:seq}
\end{minipage}
\begin{minipage}{.49\linewidth}
{\scriptsize
\begin{lstlisting}
split(Iter[T]) $\rightarrow$ (Iter[T], Iter[T])
union(List[Iter[T]],
      weights: List[float]) $\rightarrow$ Iter[T]
union_async(List[Iter[T]]): Iter[T]
\end{lstlisting}
}
\centering
\includegraphics[width=0.98\linewidth]{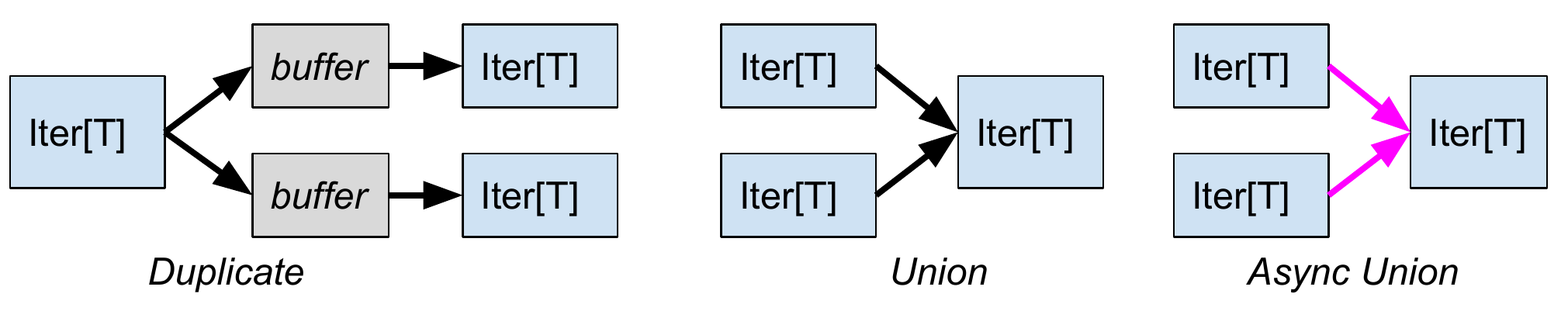}
\caption{Concurrency}
\label{fig:concurrency}
\end{minipage}
\end{figure}
\textbf{Sequencing}: To consume a parallel iterator, the items have to be serialized into some sequential order. This is the role of sequencing operators. Once converted into a sequential iterator, \texttt{next} can be called on the iterator to fetch a concrete item from the iterator. The \texttt{gather\_async} operator is used in A3C, and gathers computed gradients as fast as they are computed for application to a central policy. For a deterministic variation, we could have instead used \texttt{gather\_sync}, which waits for one gradient from each shard of the iterator before returning. The sync gather operator also has \textit{barrier semantics}. Upstream operators connected by a synchronous dependencies (black arrows) are fully halted between item fetches. This allows for the source actors to be updated prior to the next item fetch.
Barrier semantics do not apply across asynchronous dependencies, allowing the mixing of synchronous and async dataflow fragments separated by pink arrows, in~\fig{seq}.
%As in Figure \ref{fig:a3c}, we show asynchronous dependencies as pink arrows, and sync dependencies as black arrows. 

% \input{figTex/implemenations}
\textbf{Concurrency}: Complex algorithms may involve multiple concurrently executing dataflow fragments. %For example, there may be a flow producing and storing rollouts into a distributed replay buffer, and another flow sampling rollouts from the buffer and updating the policy.
Concurrency (\texttt{union}) operators, in~\fig{concurrency}, govern how these concurrent iterators relate to each other. For example, one may wish two iterators to execute sequentially in a round robin manner, execute independently in parallel, or rate limiting progress to a fixed ratio  \cite{hoffman2020acme}. %Rate limiting is identified as a key property for RL systems in Reverb \cite{acme}.
Additionally, one might wish to duplicate (\texttt{split}) an iterator, in which case buffers are automatically inserted to retain items until fully consumed. In this case, the \flowname scheduler tries to bound memory usage by prioritizing the consumer that is falling behind.

%\textbf{Scheduling and Nesting}: In our implementation of \flowname, scheduling is static. Parallel transformations are executed on the nearest upstream source actor. Sequential transformations are executed on the process calling \texttt{next} on the iterator. \flowname iterators can be nested through the underlying actor model: a new parallel iterator can be produced from a set of actors each consuming an independent iterator. This enables nested or tree-like parallel topologies.

\section{Implementation}
\label{sec:implementation}

% \zhanghao{TODO: The algorithm implemented in the original RLlib vs. the new version side by side -- "In our de-anonymized paper we will link to real before/after algorithm code on GitHub for all the entries in Table 1."}

% \zhanghao{4. highlight the differences with Ray in the revision. Ray is an actor framework. AnonLib uses Ray primitives to define execution plans for RL algorithms.}

We implemented \flowname on
%a well-known distributed Python actor framework
the Ray distributed actor framework \cite{moritz2018ray}
as two separate modules: a general purpose parallel iterator library (1241 lines of code), and a collection of RL specific dataflow operators (1118 lines of code) (\fig{arch}). We then ported the full suite of 20+ RL algorithms in \libname to \flowname, replacing the original implementations built directly on top of low-level actor and RPC primitives.
Only the portions of code in \libname related to distributed execution were changed (the exact same numerical computations are run in our port), which allows us to fairly evaluate against it as a baseline. In this section we overview two simple examples to illustrate \flowname. MAML case study, can be found in~\sect{more_implementations}.

\subsection{Asynchronous Optimization in \flowname vs \libname}
As previously seen in \fig{a3c}, A3C is straightforward to express in \flowname. \code{a3c-code} shows pseudocode for A3C in \flowname (11 lines), which we compare to a simplified version of the \libname implementation (originally 87 lines). %and \code{a2c-code} shows the synchronous A2C variation. Note that several operators take references to actors in their constructor (e.g., \texttt{ParallelRollouts}, \texttt{ReportMetrics}). This allows these operators to send messages to these actors during operator execution.
% \lstinputlisting[language=Python, caption=A3C Pseudocode., label=alg:a3c-code]{snippets/a3c.py}
% \listspace
% \begin{listing}[h]
% \inputminted{python}{snippets/a3c.py}
% \vspace{-15pt}
% \caption{A3C Pseudocode}
% \label{alg:a3c-code}
% \end{listing}
%The main difference between A3C and A2C is use of the \texttt{gather\_async} vs \texttt{gather\_sync}. In addition, A3C computes gradients as a parallel transformation prior to sequencing, whereas A2C computes and applies gradients internally in the \texttt{TrainOneStep} operator.
% \label{sect:impl-comp}
%We also compare the pseudocode of A3C implemented in the original \libname in~\code{a3c-rllib-code}. Note that we only show a small portion (model update part) of the original RLlib code for space reasons, and the real RLlib code is actually much longer and more complicated. However the \flowname code, including sample collection, model updates, etc, is pretty much as written in~\code{a3c-code}. All the codes in \code{a3c-rllib-code} are roughly equivalent to the 4 lines in~\flowname (Line 6-9 in~\code{a3c-code}).
\flowname hides the low-level worker management and data communication with its dataflow operators, providing more readable and flexible code. More detailed comparison of implementations in~\flowname and RLlib can be found in \sect{code-comparison}.
% More examples could be found on GitHub\footnote{\scriptsize Orignal: \url{https://github.com/ray-project/ray/tree/releases/0.7.7/rllib}}~\footnote{\scriptsize \flowname-based: \url{https://github.com/ray-project/ray/tree/releases/1.0.0/rllib}}. 
\begin{figure}[h]
% https://docs.google.com/document/d/16CPp7yJbWqI1PbHlcvARNjxqZVGrMyd0l3GZ4zUsiFA/edit?usp=sharing
\centering
\begin{subfigure}[b]{0.48\linewidth}
\vspace{-10pt}
    \centering
    %\includegraphics[width=.65\linewidth]{figures%/code/a2c-code.pdf}
  %  \vspace{-2pt}
 %   \caption{A2C in \flowname.}
%    \label{alg:a2c-code}
    \includegraphics[width=0.9\linewidth]{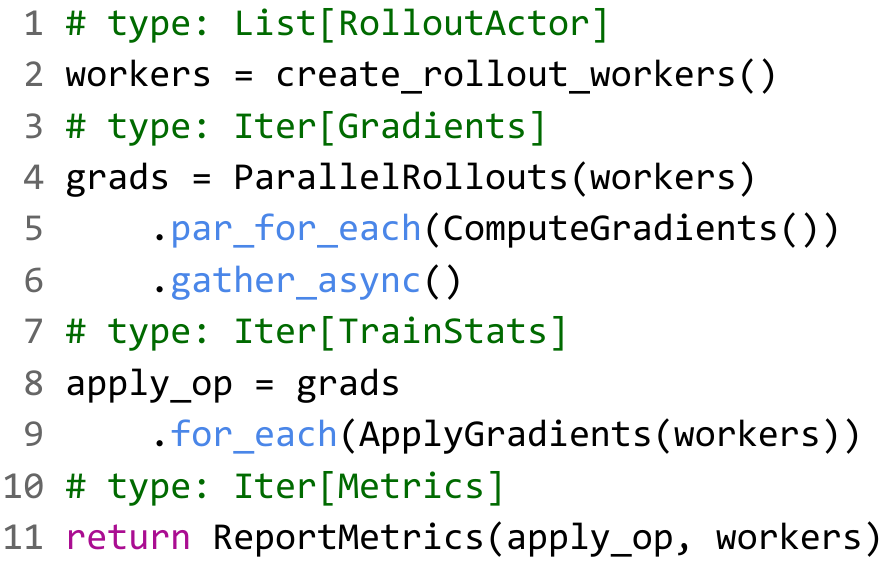}
    \vspace{45pt}
    \caption{The entire A3C dataflow in \flowname.}
    \label{alg:a3c-code}
\end{subfigure}
\begin{subfigure}[b]{0.48\linewidth}
    \centering
    \includegraphics[width=0.8\linewidth]{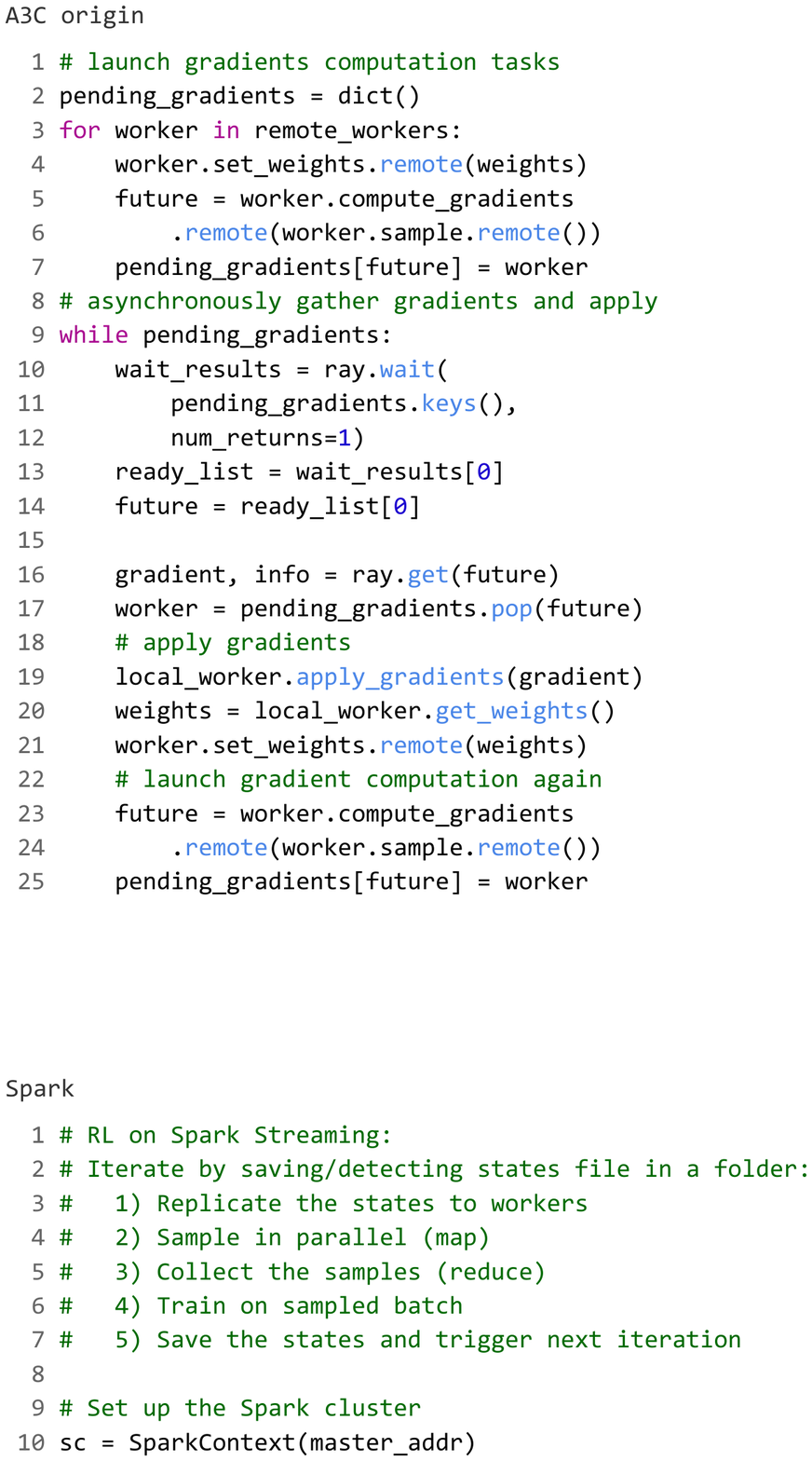}
    \vspace{-2pt}
    \caption{A small portion of the RLlib A3C policy optimizer.}
    \label{alg:a3c-rllib-code}
\end{subfigure}
% \figspace
% \vspace{-5pt}
\caption{Comparing the implementation of asynchronous optimization in \flowname vs \libname.}
\label{alg:a3c-code-compare}
% \vspace{-10pt}
\end{figure}

\subsection{Ape-X Prioritized Experience Replay in \flowname}
\begin{figure*}[h]
% https://docs.google.com/document/d/16CPp7yJbWqI1PbHlcvARNjxqZVGrMyd0l3GZ4zUsiFA/edit?usp=sharing
\centering
% \begin{subfigure}[b]{0.39\linewidth}
%     % \centering
%     \includegraphics[width=.8\linewidth]{figures/code/a3c-code.pdf}
%     \vspace{-6pt}
%     \caption{A3C}
%     \label{alg:a3c-code}
% \end{subfigure}
% ~
% \begin{subfigure}[b]{0.31\linewidth}
%     \includegraphics[width=\linewidth]{figures/code/a2c-code.pdf}
%     \vspace{-6pt}
%     \caption{A2C}
%     \label{alg:a2c-code}
% \end{subfigure}
% ~
\begin{subfigure}[b]{0.48\linewidth}
    \includegraphics[width=0.98\linewidth]{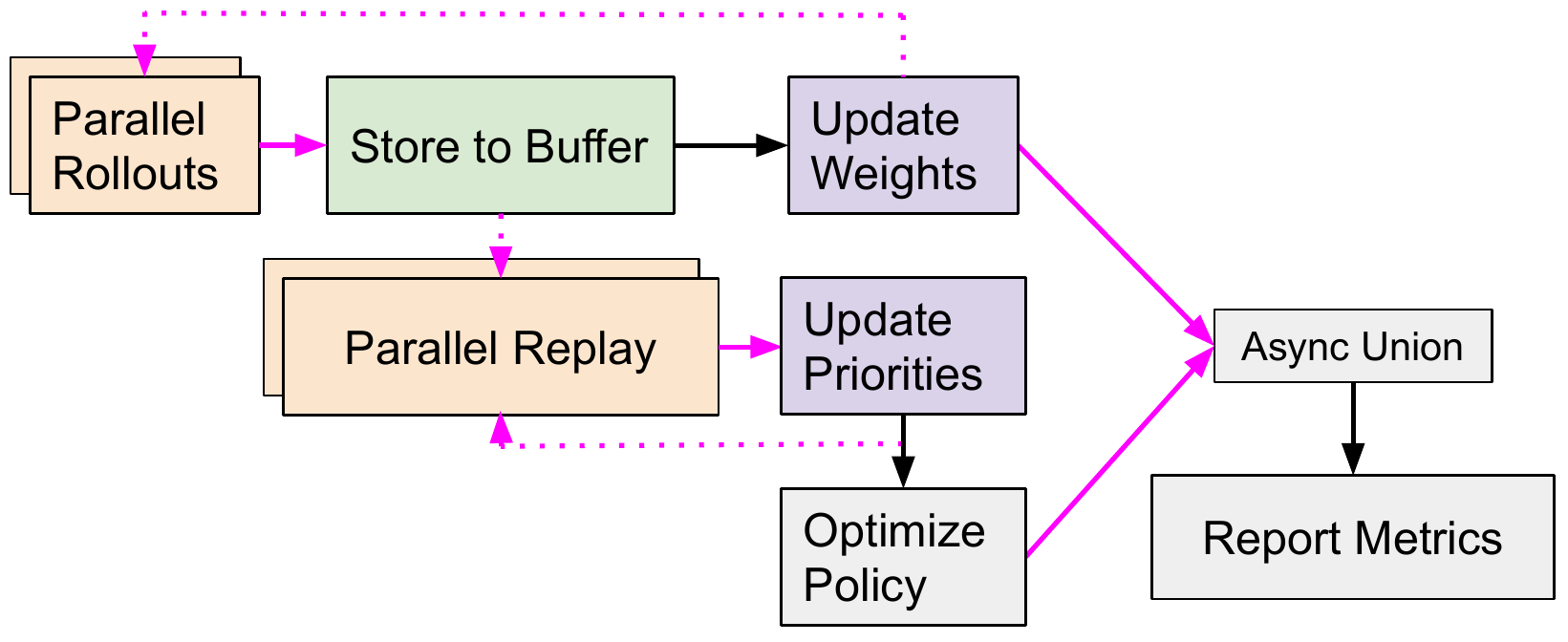}
    \vspace{38pt}
\caption{Ape-X Dataflow diagram.}
\label{fig:apex}
\end{subfigure}
~
\begin{subfigure}[b]{0.48\linewidth}
    % \centering
    \includegraphics[width=\linewidth]{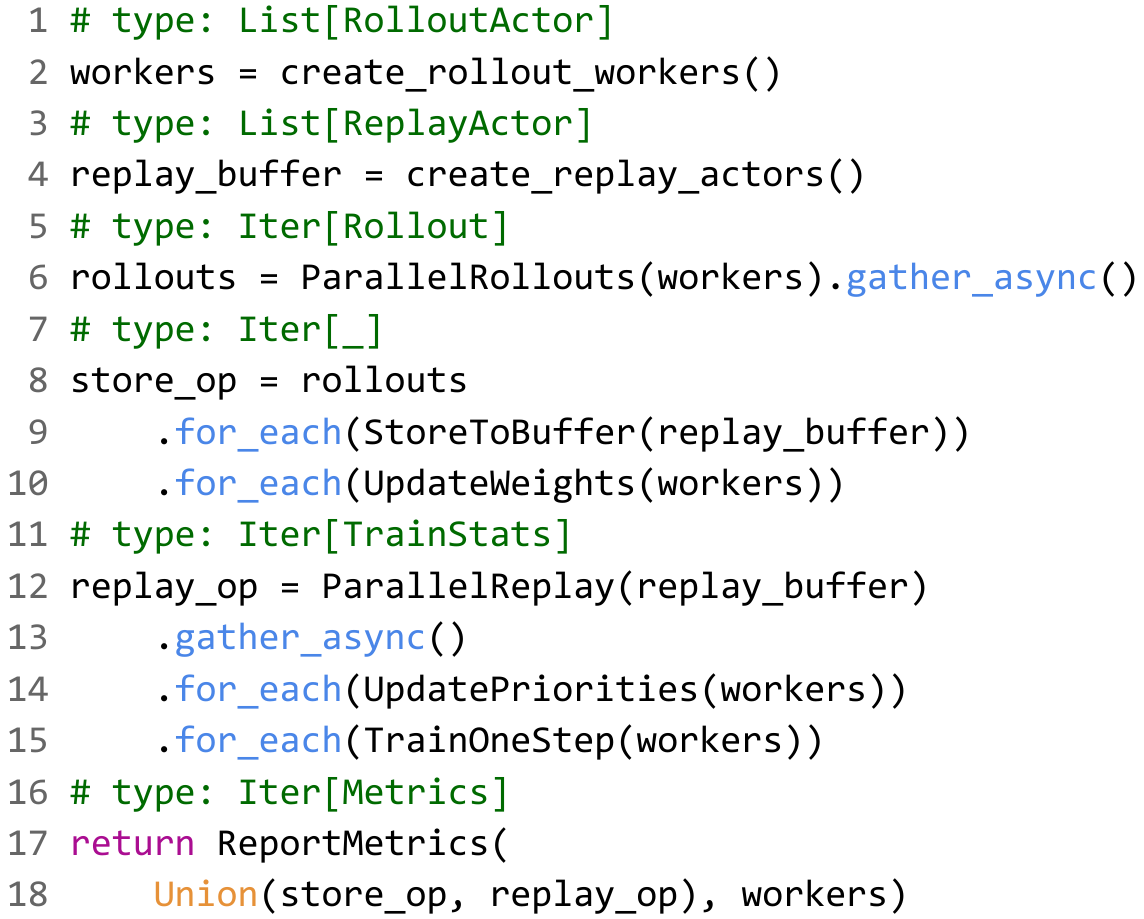}
    % \vspace{-6pt}
    \caption{Ape-X implementation in~\flowname}
    \label{alg:apex-code}
\end{subfigure}
\caption{Dataflow and implementation for Ape-X algorithm. Two dataflow fragments are executed concurrently to optimize the policy.}
\end{figure*}

Ape-X \cite{apex} (\fig{apex}) is a high-throughput variation of DQN. It is notable since it involves multiple concurrent sub-flows (experience storage, experience replay), sets of actors (rollout actors, replay actors), and actor messages (updating model weights, updating replay buffer priorities). 
The sub-flows ($\texttt{store\_op}$, $\texttt{replay\_op}$) can be composed in \flowname as follows using the $\texttt{Union}$ operator (\code{apex-code}). \revise{The complicated workflow can be implemented in several lines, as shown in~\code{apex-code}.}

\subsection{Composing DQN and PPO in Multi-Agent Training}
\label{sec:twotrainer}
\begin{figure}[h]
\centering
\begin{subfigure}[b]{0.59\linewidth}
\includegraphics[width=.95\linewidth]{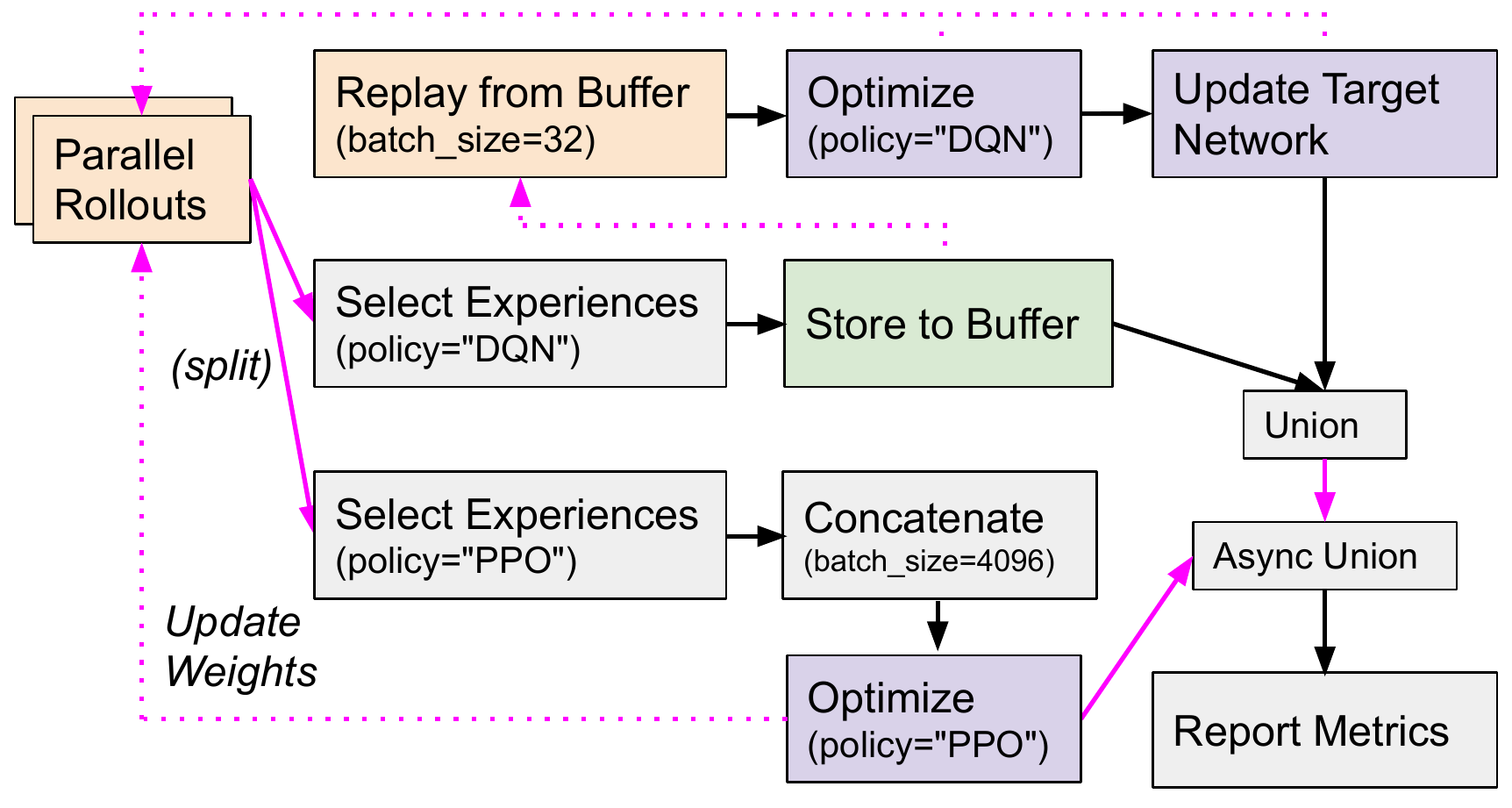}
% \vspace{-5pt}
\caption{Dataflow diagram.}
\label{fig:twotrainer}
\end{subfigure}
~
\begin{subfigure}[b]{.39\linewidth}
\centering
    \includegraphics[width=1\linewidth]{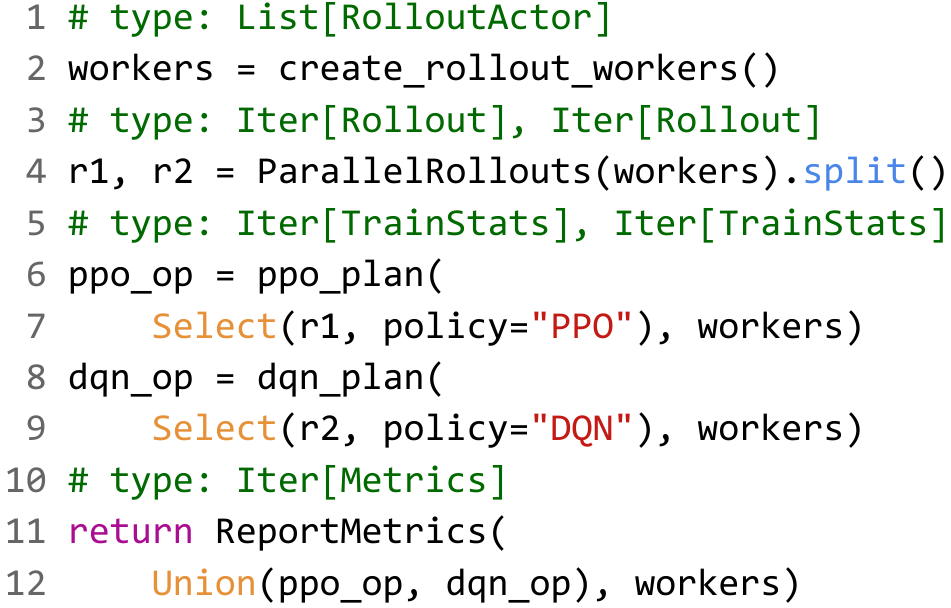}
    % \vspace{-5pt}
    \caption{Implementation in \flowname.}
    \label{alg:twotrain-code}
\end{subfigure}
\caption{Dataflow and implementation for concurrent multi-agent multi-policy workflow with PPO and DQN agents in an environment.}
% \vspace{-15pt}
\end{figure}

Multi-agent training can involve the composition of different training algorithms (i.e., PPO and DQN). \fig{twotrainer} shows the combined dataflow for an experiment that uses DQN to train certain policies in an environment and PPO to train others. \revise{The code can be found in~\code{twotrain-code}}. 
In an actor or RPC-based programming model, this type of composition is difficult because dataflow and control flow logic is intermixed. However, it is easy to express in \flowname using the \texttt{Union} operator \revise{(\fig{concurrency})}. 
In~\code{sub-flow}, we show the implementation of the two subflow, \texttt{ppo\_plan} and \texttt{dqn\_plan}, in the multi-agent training (\code{twotrain-code}). 

\begin{figure*}[h]
% https://docs.google.com/document/d/16CPp7yJbWqI1PbHlcvARNjxqZVGrMyd0l3GZ4zUsiFA/edit?usp=sharing
% \centering
\begin{subfigure}[b]{0.48\linewidth}
    \includegraphics[width=1.\linewidth]{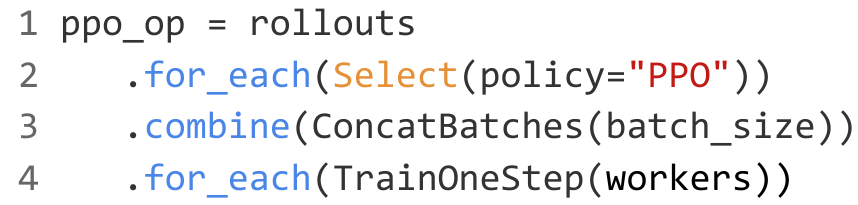}
    \vspace{6pt}
\caption{Implementation of PPO Subflow.}
\label{fig:ppo-subflow}
\end{subfigure}
\begin{subfigure}[b]{0.48\linewidth}
    % \centering
    \includegraphics[width=1.\linewidth]{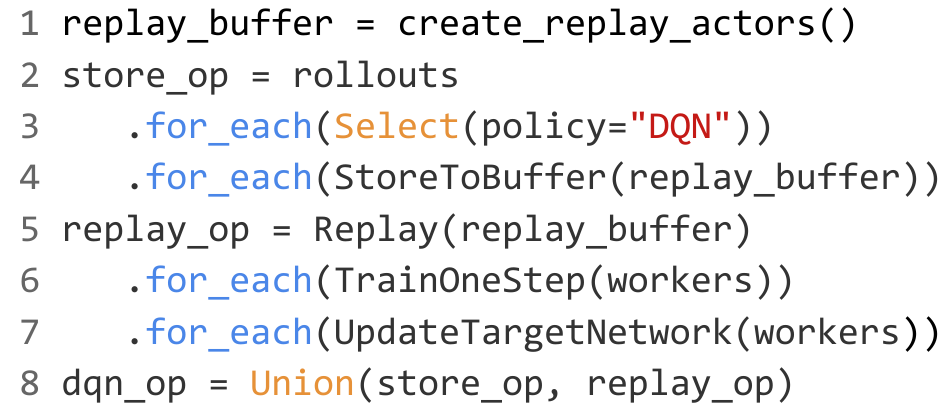}
    \caption{Implementation of DQN Subflow.}
    \label{alg:dqn-subflow}
\end{subfigure}
\caption{Implementation of sub-flows for the multi-agent multi-policy training of PPO and DQN.}
\label{alg:sub-flow}
\end{figure*}
% \lstinputlisting[language=Python, caption=Twin Trainer Pseudocode., label=alg:twotrain-code]{snippets/twotrain.py}
% \listspace

% \input{figTex/twotrainer}
% \begin{listing}[t]
% \inputminted{python}{snippets/twotrain.py}
% \vspace{-15pt}
% \caption{Twin Trainer Pseudocode}
% \label{alg:twotrain-code}
% \end{listing}
\section{Evaluation}
\label{sec:evaluation}

In our evaluation, we seek to answer the following questions:

\begin{enumerate}
[leftmargin=*]
\item What is the quantitative improvement in code complexity with \flowname?
\item How does \flowname compare to other systems in terms of flexibility and performance for RL tasks?
\end{enumerate}
%\setlist{nosep}
%\begin{itemize}[leftmargin=*,noitemsep]
%  \item % We examine the total lines of code and modularity of implementations.
 % \item Can \flowname execute RL workloads with high performance? %We address this with microbenchmarks and performance comparisons across implementations.
  %\item % We compare against Spark Streaming and analyze its limitations.
%\end{itemize}

\subsection{Code Complexity}

% \revise{%
%\flowname is a high-level API that simplifies the code for distributed RL algorithms. In Appendix A.1, we provide a motivating example of how the A3C algorithm is implemented in the original \libname, compared to the new \flowname-based implementation.
%The execution logic of the original implementation is highly combined with the low-level RPCs, leading to a less readable and flexible implementation with more lines of codes. 
% \lstinputlisting[language=Python, caption={A3C Pseudocode in original \libname}, label=alg:a3c-origin-code]{snippets/a3c-origin.py}
% \listspace
% }

\textbf{Lines of Code}:
In \tab{loc-algo} we compare the original algorithms in \libname to after porting to \flowname. No functionality was lost in the \flowname re-implementations. We count all lines of code directly related to distributed execution, including comments and instrumentation code, but not including utility functions shared across all algorithms. For completeness, for \flowname we include both an minimal (\texttt{\flowname}) and conservative (\texttt{+shared}) estimate of lines of code. The conservative estimate includes lines of code in shared operators. Overall, we observe between a 1.9-9.6$\times$ (optimistic) and 1.1-3.1$\times$ (conservative) reduction in lines of code with \flowname. The most complex algorithm (IMPALA) shrunk from 694 to 89-362 lines.% To motivate the simplicity, we also compare the implementation of A3C algorithm based on \flowname and RLlib in \code{a3c-code-compare}.
\begin{table}[h]
\centering
\caption{Lines of code for several prototypical algorithms implemented with the original \libname vs our \flowname-based \libname.
\textit{*Original MAML: \url{https://github.com/jonasrothfuss/ProMP}}}
\vspace{5pt}
{\small
\begin{tabular}{lcccr}
    \toprule
            & RLlib & \flowname & +shared & Ratio  \\
    \midrule
    A3C & 87 & 11 & 52 & 1.6-9.6$\times$ \\  % + async grads
    A2C & 154 & 25 & 50 & 3.1-6.1$\times$ \\ % + avg grad
    DQN & 239 & 87 & 139 & 1.7-2.7$\times$\\  % + update target
    PPO & 386 & 79 & 225 & 1.7-4.8$\times$\\  % + train multi gpu
    Ape-X & 250 & 126 & 216 & 1.1-1.9$\times$ \\  % + learner thread
    IMPALA & 694 & 89 & 362 & 1.9-7.8$\times$ \\  % + tree agg, multi gpu
    MAML & 370* & 136 & 136 & 2.7$\times$ \\ 
    \bottomrule
\end{tabular}
}
% \tabspace
\vspace{-10pt}
\label{tab:loc-algo}
\end{table}

\textbf{Readability}: We believe \flowname provides several key benefits for readability of RL algorithms:
\begin{enumerate}[leftmargin=*]

\item The high-level dataflow of an algorithm is visible at a glance in very few lines of code, allowing readers to understand and modify execution pattern without diving deep into the execution logic.

\item Execution logic is organized into individual operators, each of which has a consistent input and output interface (i.e., transforms an iterator into another iterator). In contrast to building on low-level RPC systems, developers can decompose their algorithms into reusable operators.

\item Performance concerns are isolated into the lower-level parallel iterator library. Developers do not need to deal with low-level concepts such as batching or flow-control.

\end{enumerate}

\textbf{Flexibility}: As evidence of \flowname's flexibility, an undergraduate was able to implement several model-based (e.g., MB-MPO) and meta-learning algorithms (e.g., MAML), neither of which fit into previously existing execution patterns in \flowname. This was only possible due to the flexibility of \flowname's model. 
% that mixes together the actor and dataflow model. 
% \code{maml-code} concisely expresses MAML's dataflow (also shown in Figure \ref{fig:maml}) \cite{finn2017modelagnostic}. The MAML dataflow involves nested optimization loops; workers collect pre-adaptation data, perform inner adaptation (i.e., individual optimization calls to an ensemble of models spread across the workers), and collect post-adaptation data. Once inner adaptation is complete, the accumulated data is batched together to compute the meta-update step, which is broadcast to all workers. 
\flowname captures MAML in 139 lines compared to a baseline of $\approx$370 lines (\tab{loc-algo}). Detailed discussion can be found in~\sect{maml}.
%In our code snippet, we modularize MAML-PPO \cite{rothfuss2018promp} by creating two classes: \texttt{inner\_adaptation\_steps} and \texttt{MetaUpdate}. The class \texttt{inner\_adaptation\_steps} accumulates worker-generated data and computes inner adaptation for workers while \texttt{MetaUpdate} performs meta-update step and returns logging statistics.

% \input{figTex/maml-code}
% \lstinputlisting[language=Python, caption={MAML Pseudocode.}, label=alg:maml-code]{snippets/maml.py}
% \listspace
% \begin{listing}[h]
% \inputminted{python}{snippets/maml.py}
% \vspace{-15pt}
% \caption{MAML Pseudocode}
% \label{alg:maml-code}
% \end{listing}
% \input{figTex/maml-code}

% \zhanghao{Propagation cost: measures the extent to which a change in one element impacts other elements. It is a representation of the degree of coupling without consideration of the proximity between elements}

\subsection{Microbenchmarks and Performance Comparisons}
% \zhanghao{TODO: "emphasize that the goal of AnonFlow is not to improve the performance of RL algorithms, but to make it possible to, in the same framework, make complex RL training use cases accessible to RL researchers and practitioners"}
%\revise{\flowname aims to make complex RL training use cases accessible to RL practitioners, without introduce performance overheads. In this section, we will show that \flowname has a similar performance compared to the original \libname.}
For all the experiments, we use a cluster with an AWS p3.16xlarge GPU head instance with additional m4.16xlarge worker instances. All machines have 64 vCPUs and are connected by a 25Gbps network. More experiments for different RL algorithms can be found in \url{https://github.com/ray-project/rl-experiments}.
\begin{figure}[t]

\begin{subfigure}[t]{0.48\linewidth}
\centering
% link: https://1drv.ms/p/s!Ah3_lLUfMlckjbtmiXKWnJwHHbxXUQ?e=VMFBXI
\includegraphics[width=0.99\linewidth]{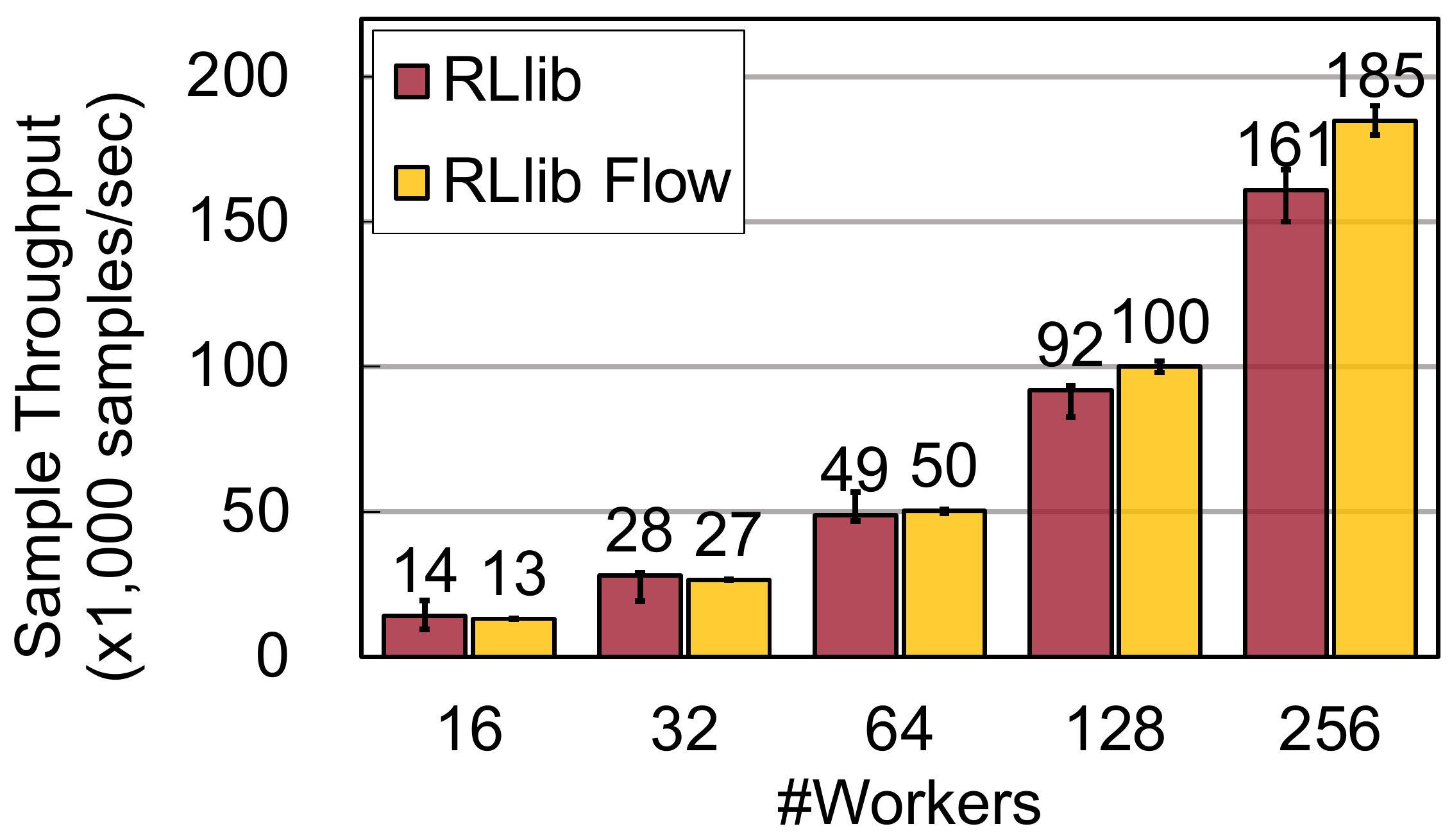}
\vspace{-15pt}
\caption{Sample efficiency on CartPole.}
% CartPole-v0 scales on AWS m4.16xlarge CPU instances with IMPALA and a dummy model.
% Our \flowname-based \libname achieves comparable throughput at small scale and slightly better (1.1\x) at larger scale. The red bars indicate the min and max value of the sample throughput.
\label{fig:sample-efficiency}
\end{subfigure}
~
\begin{subfigure}[t]{0.48\linewidth}
\centering
    % link: https://1drv.ms/p/s!Ah3_lLUfMlckjbtqH49JqfLAko1eMQ?e=YsX2kL
    \includegraphics[width=0.99\linewidth]{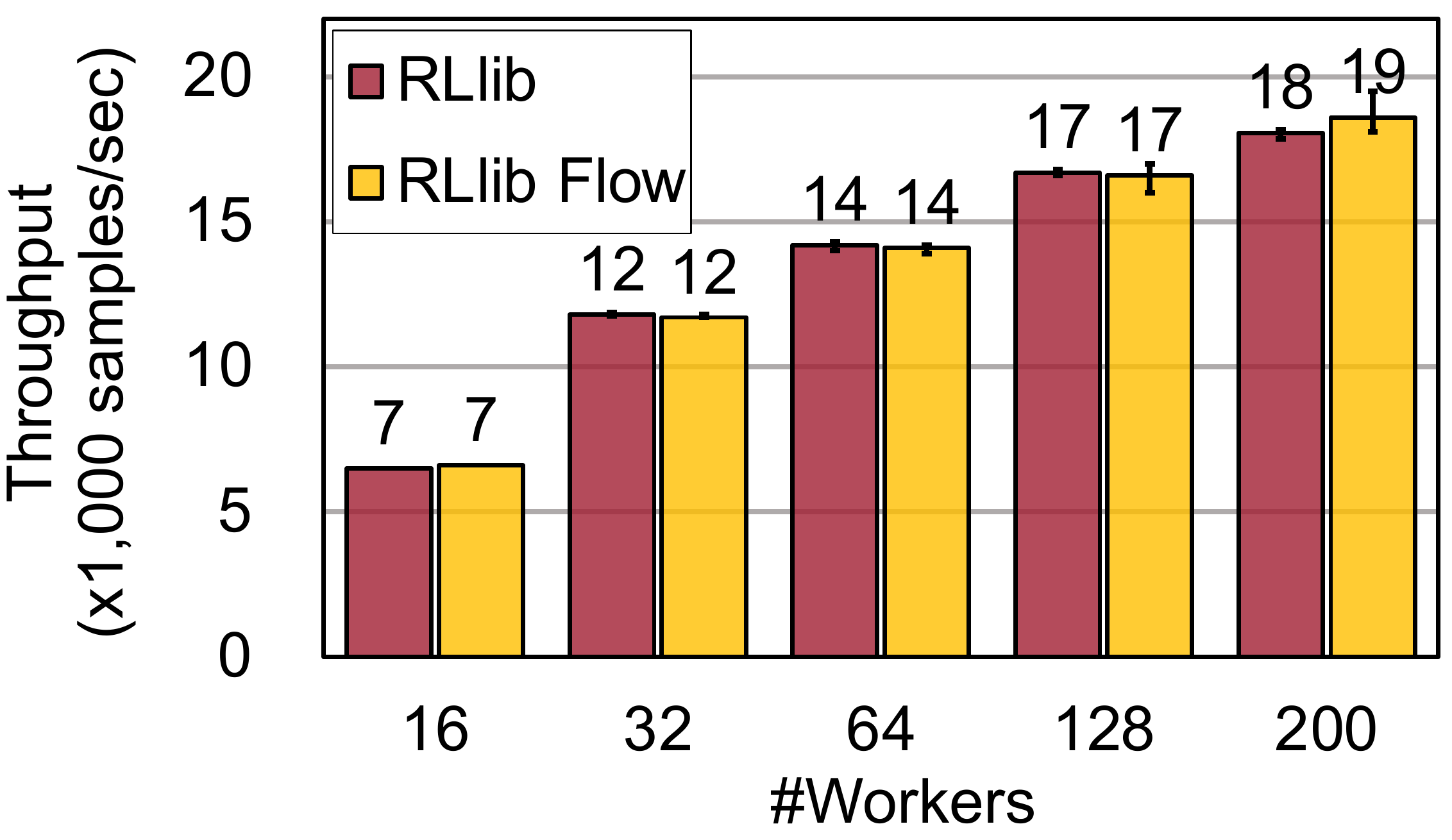}
\vspace{-15pt}
\caption{Training throughput on Atari.}
% The experiments are conducted on PongNoFrameSkip-v4 using IMPALA algorithm with AWS p3.16xlarge as the master and additional m4.16xlarge as actor workers. We achieve comparable performance across different versions of \libname. The red bars indicate the min and max throughput of three runs.
\label{fig:throughput}
\end{subfigure}
\vspace{-5pt}
\caption{Performance of RLlib compared with \flowname, executing identical numerical code. \revise{\flowname achieves comparable or better performance across different environments.% We hypothesize that the improvements at large scale come from small optimizations such as batched RPC wait.
}}
% \vspace{-15pt}
\end{figure}

\textbf{Sampling Microbenchmark}: We evaluate the data throughput of \flowname in isolation by running RL training with a dummy policy (with only one trainable scalar).
% This enables us to observe how fast the framework can support sample collection from asynchronous environments. The SGD batch size and unroll length are fixed to \numprint{100}K and 250, respectively.
\fig{sample-efficiency} shows that \flowname achieves slightly better throughput due to small optimizations such as batched RPC wait, which are easy to implement across multiple algorithms in a common way in \flowname.

\textbf{IMPALA Throughput}: In \fig{throughput} we benchmark IMPALA, one of \libname's high-throughput RL algorithms, and show that \flowname achieves similar or better end-to-end performance.
%We measured the throughput of Atari BreakoutNoFrameSkip-v4 and Pendulum-v0 environments with 32 and 2 workers, respectively. We also test the scalability of IMPALA for PongNoFrameSkip-v4 environments using various number of workers.% The number of the CPU instance is set that the total vCPUs are exactly enough for the number of workers.
%\input{figTex/throughput-tab}
%\fig{throughput}%and \tab{throughput-tab}
%indicates that our new \flowname-based \libname achieves comparable throughput to the original \libname on various environments. In \fig{throughput}, as the number of nodes increases, the new \libname also shows similar scalability as the original one. That shows that \flowname does not impose overheads compared to a low-level implementation in a realistic scenario.
% \input{figTex/throughput-pong}

\begin{figure}[t]
\begin{minipage}[b]{.49\linewidth}
\centering
% link: https://1drv.ms/p/s!Ah3_lLUfMlckjbtfFb8iw4nQmc6qtg?e=PMlppD
\includegraphics[width=\linewidth]{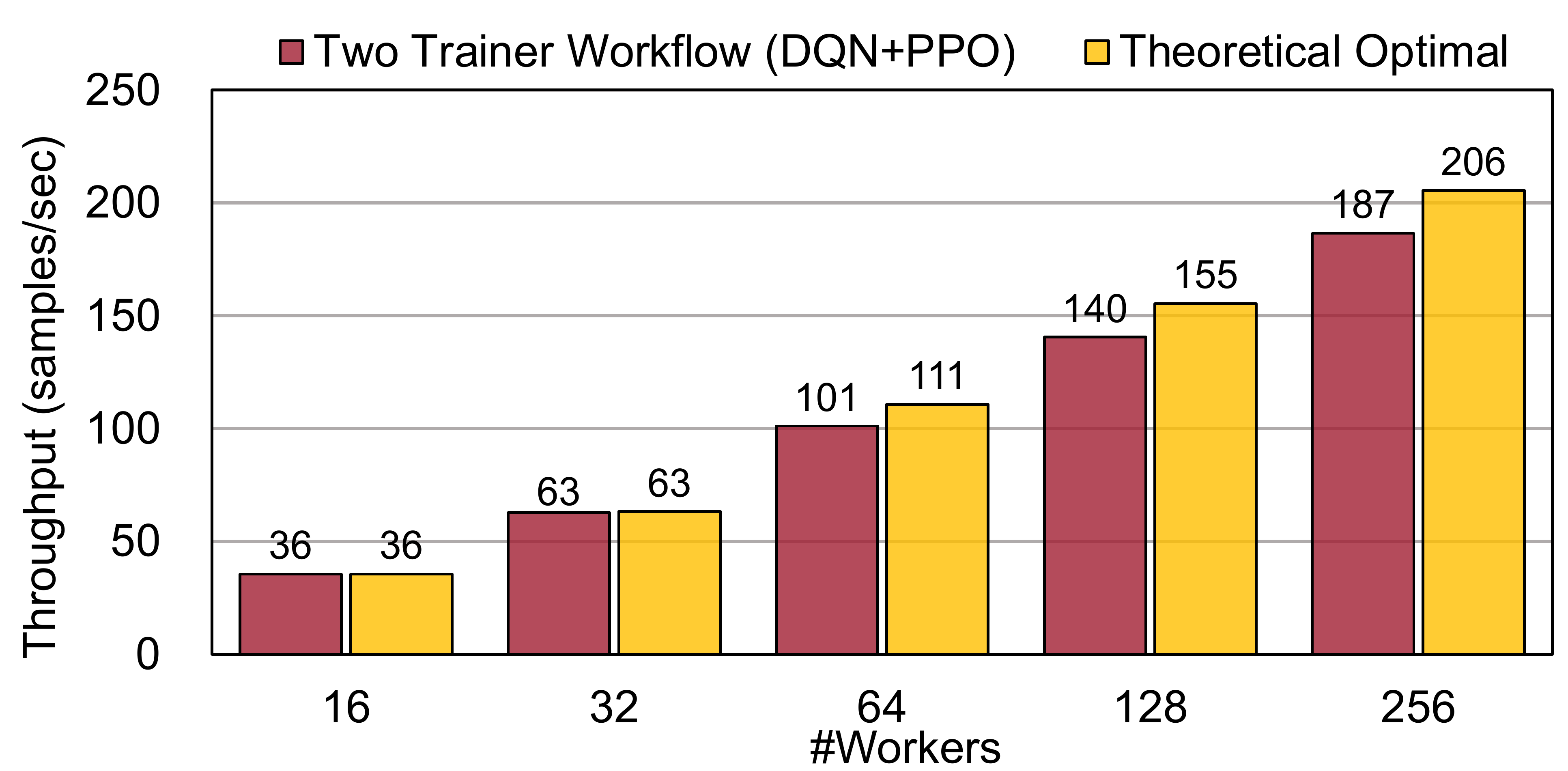}
% \vspace{-15pt}
\caption{\flowname achieves close to the theoretical optimal performance combining two workflows for multi-agent training, making it practical to use for composing complex training scenarios.}
% The running time is measured for one iteration on AWS m4.10xlarge CPU instances with 40 vCPUs and 10Gbps network each. \flowname achieves better training throughput and scalability against a Spark Streaming based implementation with up to 2.9\x faster speed.
\label{fig:rllib-performance}
\end{minipage}
~
\begin{minipage}[b]{.49\linewidth}
% https://docs.google.com/drawings/d/11pJ7KkblxeipjecAqdBOk-iZUZid43dItsFt3Bd8TYA/edit
\centering
% link: https://1drv.ms/p/s!Ah3_lLUfMlckjbtfFb8iw4nQmc6qtg?e=PMlppD
\includegraphics[width=\linewidth]{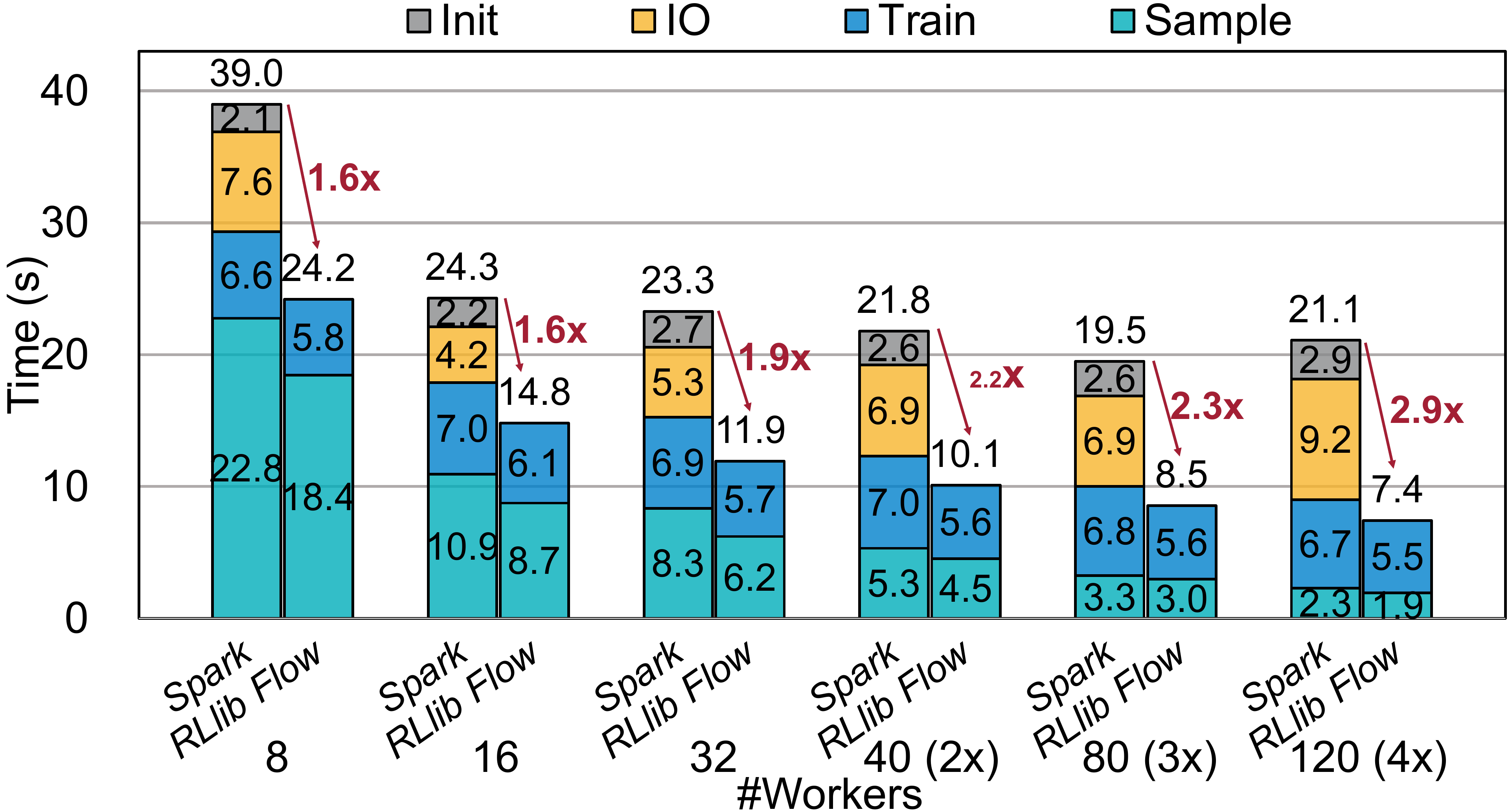}
\vspace{-10pt}
\caption{\revise{The throughput comparison between \flowname and Spark Streaming with PPO algorithm on CartPole-v0 environment.}}
% The running time is measured for one iteration on AWS m4.10xlarge CPU instances with 40 vCPUs and 10Gbps network each. \flowname achieves better training throughput and scalability against a Spark Streaming based implementation with up to 2.9\x faster speed.
\label{fig:rllib-spark}
\end{minipage}
% \vspace{-25pt}
\end{figure}

\textbf{Performance of Multi-Agent Multi-Policy Workflow}:
In \fig{rllib-performance}, we show that the workflow of the two-trainer example (\fig{twotrainer}) achieves close to the theoretical best performance possible combining the two workflows (calculated via Amdahl's law). This benchmark was run in a multi-agent Atari environment with four agents per policy, and shows \flowname can be practically used to compose complex training workflows.

%With the power of \texttt{Union} operator (\fig{concurrency}), people can easily combine different workflows to form previously unsupported algorithms such as the two trainer workflow in~\sect{twotrainer} and~\fig{twotrainer}. In~\fig{rllib-performance}, we show the scalability of the workflow in \flowname on the Atari BreakoutNoFrameSkip-v4 environment with 4 agents each policy. We also calculate the upper bound of the theoretical best throughput, according to the amdahl's law based on the throughput of DQN and PPO subtasks. The negligible gap between our \flowname and the theoretical results indicates that users can easily compose multiple workflows in our \flowname without losing efficiency.
% for the two-trainer workflow with Amdahl's law in~\eqn{amdahl}.
% \begin{equation}
% \label{eqn:amdahl}
%     f(s_\text{Two Trainer}) = \nicefrac{1}{\frac{p_\text{PPO}}{s_\text{PPO}}+ \frac{p_\text{DQN}}{s_\text{DQN}}+\frac{p_\text{remaining}}{s_\text{remaining}}} \leq \nicefrac{1}{\frac{p_\text{PPO}}{s_\text{PPO}}+ \frac{p_\text{DQN}}{s_\text{DQN}}+\frac{p_\text{remaining}}{r_\text{workers}}}
% \end{equation}
% where $f$ is the theoretical speed up; $p_i$ and $s_i$ are the portion and the speed up of the subtask $i$, respectively, and $r_\text{workers}$ is the ratio of the number of workers after and before scaling up.
% We show the optimal throughput as the error bars on the bars for the two trainer workflow. It indicates that our \flowname could combine two dataflow efficiently, with negligible gap to the upper bound of the optimal throughput. 

\textbf{Comparison to Spark Streaming}:
\label{sec:spark-streaming}
Distributed dataflow systems such as Spark Streaming~\cite{zaharia_spark_stream_2013} and Flink~\cite{Carbone2015ApacheFS} are designed for collecting and transforming live data streams from online applications (e.g., event streams, social media). Given the basic \textit{map} and \textit{reduce} operations, we can implement synchronous RL algorithms in any of these streaming frameworks. However, without consideration for the requirements of RL tasks (Section \ref{sec:streaming}), these frameworks can introduce significant overheads. In \fig{rllib-spark} we compare the performance of PPO implemented in Spark Streaming and \flowname. Implementation details are in in \apdx{spark}. 

\section{Related Work}
\label{Related Work}

\textbf{Reinforcement Learning Systems}: \flowname is implemented concretely in \libname, however, we hope it can provide inspiration for a new generation of general purpose RL systems.
RL libraries available today range from single-threaded implementations \cite{duan2016benchmarking, baselines, castro18dopamine, tensorforce} to distributed \cite{Schaarschmidt2019rlgraph, loon2019slm, coach, liang2018rllib, falcon2019pytorch, hoffman2020acme}. These libraries often focus on providing common frameworks for the numerical concerns of RL algorithms (e.g., loss, exploration, and optimization steps).

However, these aforementioned libraries rely on \textit{predefined} distributed execution patterns.
For example, for the Ape-X dataflow in~\fig{apex}, RLlib defines this with a fixed ``AsyncReplayOptimizer''\footnote{\scriptsize\url{https://docs.ray.io/en/releases-0.7.7/\_modules/ray/rllib/optimizers/async\_replay\_optimizer.html}} class that implements the topology\revise{, intermixing the dataflow and the control flow.}; 
RLGraph uses an adapted implementation\footnote{\scriptsize\url{https://github.com/rlgraph/rlgraph/blob/master/rlgraph/execution/ray/apex/apex\_executor.py}} from RLlib as part of their Ape-X algorithm meta-graph, 
while Coach does not support Ape-X\footnote{\scriptsize\url{https://github.com/IntelLabs/coach}}.
These execution patterns are predefined as they are low-level, complex to implement, and cannot be modified using high-level end-user APIs. 
In contrast, \flowname proposes a high-level distributed programming model for RL algorithm implementation, exposing this pattern in much fewer lines of code (\code{apex-code}), and allowing free composition of these patterns by users (\code{twotrain-code}).
The ideas from \flowname can be integrated with any RL library to enable flexibility in distributed execution.

\textbf{Distributed Computation Models}: \flowname draws inspiration from both streaming dataflow and actor-based programming models. Popular open source implementations of streaming dataflow, including Apache Storm~\cite{toshniwal2014storm}, Apache Flink~\cite{Carbone2015ApacheFS}, and Apache Spark~\cite{Zaharia_apache_spark,zaharia_spark_stream_2013} transparently distribute data to multiple processors in the background, hiding the scheduling and message passing for distribution from programmers.
In~\apdx{spark}, we show how distributed PPO can be implemented in Apache Spark. Apache Flink's \texttt{Delta Iterate} operator can similarly support synchronous RL algorithms. However, data processing frameworks have limited asynchronous iteration support.

%With the function, we can place the environment simulators in the step function, which will iteratively generate trajectories in the \texttt{workset}. The trainer for the model can be placed in the \texttt{Solution Set}. It consumes the trajectories produced by the step function and updates the model weight within the simulators. In Apache Flink, the step function supports arbitrary dataflow operators, so the environment simulators can be easily distributed with map and reduce. However, the asynchronous between the trainer and the simulators are limited.%}
% \zhanghao{TODO: mention about support of map, reduce in the step function.}
% \revise{
% However, without considering RL specific characters, such as less strict consistency requirement and asynchronous message passing between stages, overheads will be introduced.
% }

The Volcano model~\cite{volcano}, commonly used for distributed data processing, pioneered the parallel iterator abstraction. \flowname builds on the Volcano model to not only encapsulate parallelism, but also to encapsulate the synchronization requirements between concurrent dataflow fragments, enabling users to also leverage actor message passing.

Naiad~\cite{murray2013naiad} is a low-level distributed dataflow system that supports cyclic execution graphs and message passing. It is designed as a system for implementing higher-level programming models.
%
%It models a directed graph with nodes and logically timestamped messages flowing along edges.
%
In principle, it is possible to implement the \flowname model in Naiad. Transformation operators can be placed on the stateful vertices of the execution graph. The message passing and concurrency (\texttt{Union}) operators can be represented by calling \texttt{\textsc{SendBy}} and \texttt{\textsc{OnRecv}} interface on senders and receivers, which support asynchronous execution. \flowname's barrier semantics can be expressed with \texttt{\textsc{OnNotify}} and \texttt{\textsc{NotifyAt}}, where the former indicates all the required messages are ready, and the latter blocks execution until the notification has been received. We implemented \flowname on Ray instead of Naiad for practical reasons (e.g., Python support).

\section{Conclusion}

In summary, we propose \flowname, a hybrid actor-dataflow programming model for distributed RL. We designed \flowname to simplify the understanding, debugging, and customization of distributed RL algorithms RL developers require. \flowname provides comparable performance to reference algorithms implemented directly on low-level actor and RPC primitives, enables complex multi-agent and meta-learning use cases, and reduces the lines of code for distributed execution in a production RL library by 2-9\x. \flowname is available as part of the open source RLlib project, and we hope it can also help inform the design of future RL libraries.
%To address the challenges, \flowname, like streaming data systems, provides a small set of operator-like primitives that can be composed to express distributed RL algorithms. With quantitative results, we demonstrate that \flowname simplifies the algorithm implementation in \libname without harming the throughput and scalability, and outperforms existing dataflow systems adapted for RL tasks, by up to 2.9\x.

\section{Acknowledgement}
In addition to NSF CISE Expeditions Award CCF-1730628, this research is supported by gifts from Amazon Web Services, Ant Group, Ericsson, Facebook, Futurewei, Google, Intel, Microsoft, Scotiabank, and VMware.
% \zhanghao{TODO: If there is any other organization to acknowledge}

{\small
\bibliographystyle{ieee}
\bibliography{reference}
}

\appendix
\renewcommand{\thesection}{A.\arabic{section}}
\renewcommand{\thefigure}{A\arabic{figure}}
\renewcommand{\thelisting}{A\arabic{listing}}
\setcounter{section}{0}
\setcounter{figure}{0}

\clearpage
%\zhanghao{TODO: Appendix needs to be polished, after the submission of main paper)}
\section{RL in Spark Streaming}
\label{append:spark}
\textbf{PPO Implementation.} In \code{spark-code}, we show the high-level pseudocode of our port of the PPO algorithm to Spark Streaming. Similar to our port of \libname to \flowname, we only changed the parts of the PPO algorithm in \libname that affect distributed execution, keeping the core algorithm implementation (e.g., numerical definition of policy loss and neural networks in TensorFlow) as similar as possible for fair comparison. We made a best attempt at working around aforementioned limitations (e.g., using a \texttt{binaryRecordsStream} input source to efficiently handle looping, defining efficient serializers for neural network state, and adjusting the microbatching to emulate the \libname configuration).

\begin{figure}[h]
% https://docs.google.com/document/d/16CPp7yJbWqI1PbHlcvARNjxqZVGrMyd0l3GZ4zUsiFA/edit?usp=sharing
\centering
\includegraphics[width=0.5\linewidth]{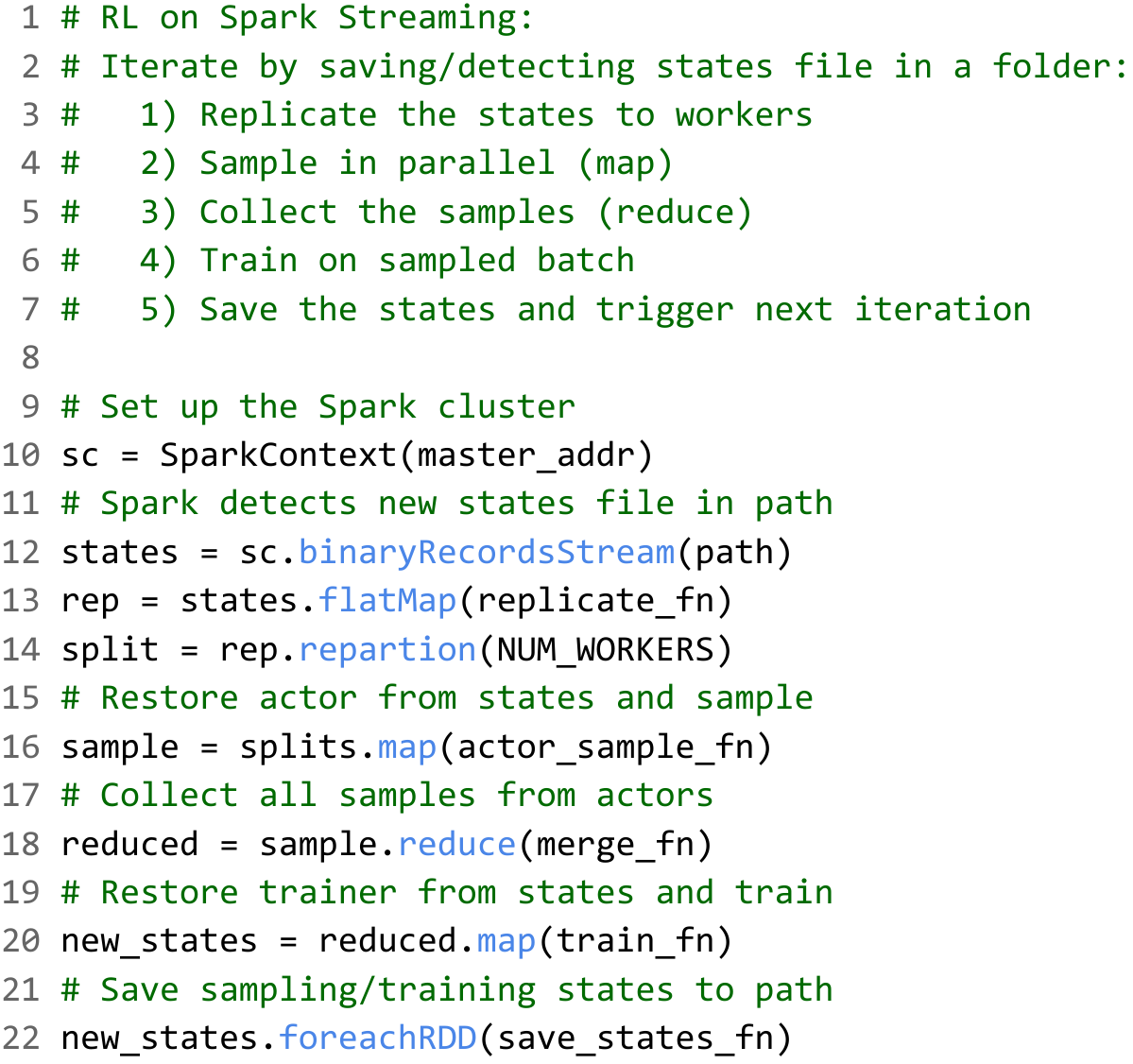}
% \figspace
\caption{Example of Spark Streaming for Distributed RL.}
\label{alg:spark-code}
% \figspace
\end{figure}
% \begin{listing}[h]
% \setminted[python]{breaklines}
% \inputminted{python}{snippets/spark.py}
% \vspace{-15pt}
% \caption{Example of Spark Streaming for Distributed RL.}
% \label{alg:spark-code}
% \end{listing}

\textbf{Experiment Setup.} We conduct comparisons between the performance of both implementations. In the experiment, we adopt the PPO algorithm for the CartPole-v0 environment with a fixed sampling batch size $B$ of 100K. Each worker samples ($B/\text{\# workers}$) samples each iteration, and for simplicity, the learner updates the model on CPU using a minibatch with 128 samples from the sampled batch. Experiments here are conducted on AWS m4.10xlarge instances.
%\fig{rllib-spark} demonstrates that \flowname PPO outperforms a Spark Streaming based implementation on RL training in both single machine and distributed scenarios. 

\textbf{Data Framework Limitations}: Spark Streaming is a data streaming framework designed for general purpose data processing. We note several challenges we encountered attempting to port RL algorithms to Spark Streaming:
\setlist{nolistsep}
\begin{enumerate}[leftmargin=*,noitemsep]
    \item Support for asynchronous operations. Data processing systems like Spark Streaming do not support asynchronous or non-deterministic operations that are needed for asynchronous RL algorithms.
    \item Looping operations are not well supported. While many dataflow models in principle support iterative algorithms, we found it necessary to work around them due to lack of language APIs (i.e., no Python API).
    \item Support for non-serializable state. In the dataflow model, there is no way to persist arbitrary state (i.e., environments, neural network models on the GPU). While necessary for fault-tolerance, the requirement for serializability impacts the performance and feasibility of many RL workloads.
    \item Lack of control over batching. We found that certain constructs such as the data batch size for on-policy algorithms are difficult to control in traditional streaming frameworks, since they are not part of the relational data processing model.
\end{enumerate}

For a single machine (the left three pairs), the breakdown of the running time indicates that the initialization and I/O overheads slow down the training process for Spark comparing to our \flowname. The former overheads come from the nature of Spark that the transformation functions do not persist variables. We have to serialize both the sampling and training states and re-initialize the variables in the next iteration to have a continuous running process. On the other hand, the I/O overheads come from looping back the states back to the input. As an event-time driven streaming system, the stream engine detects changes for the saved states from the source directory and starts new stream processing. The disk I/O leads to high overheads compared to \flowname. 

For distributed situation (the right three pairs), the improvement of \flowname becomes more significant against Spark, up to 2.9\x. As the number of workers scales up, the sampling time decreases for both the dataflow model. Still, the initialization and I/O overheads stay unchanged, leading to lesser scalability for Spark. 

\section{Implementation Examples}
\label{sec:more_implementations}

\subsection{Example: MAML}
\label{sec:maml}
\begin{figure}[h]
 %https://docs.google.com/presentation/d/10bLdyeErPKxxyCNrzhGdHG91L6F82Y_OnomxN4eIwUY/edit#slide=id.ga037b6c774_0_0
 \begin{subfigure}{.49\linewidth}
\includegraphics[width=0.98\linewidth]{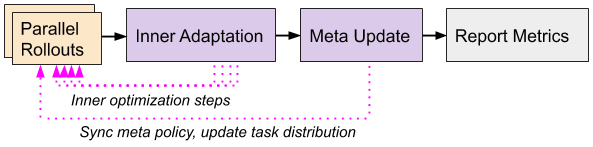}
% \figspace
\caption{MAML dataflow includes a number of nested inner adaptation steps (optimization calls to the source actors) prior to update of the meta-policy. The meta-policy update and inner adaptation steps integrate cleanly into the dataflow, their ordering guaranteed by the synchronous data dependency barrier between the inner adaptation and meta update steps.}
\label{fig:maml}
\end{subfigure}
~
\begin{subfigure}{.49\linewidth}
% https://docs.google.com/document/d/16CPp7yJbWqI1PbHlcvARNjxqZVGrMyd0l3GZ4zUsiFA/edit?usp=sharing
\centering
\includegraphics[width=0.98\linewidth]{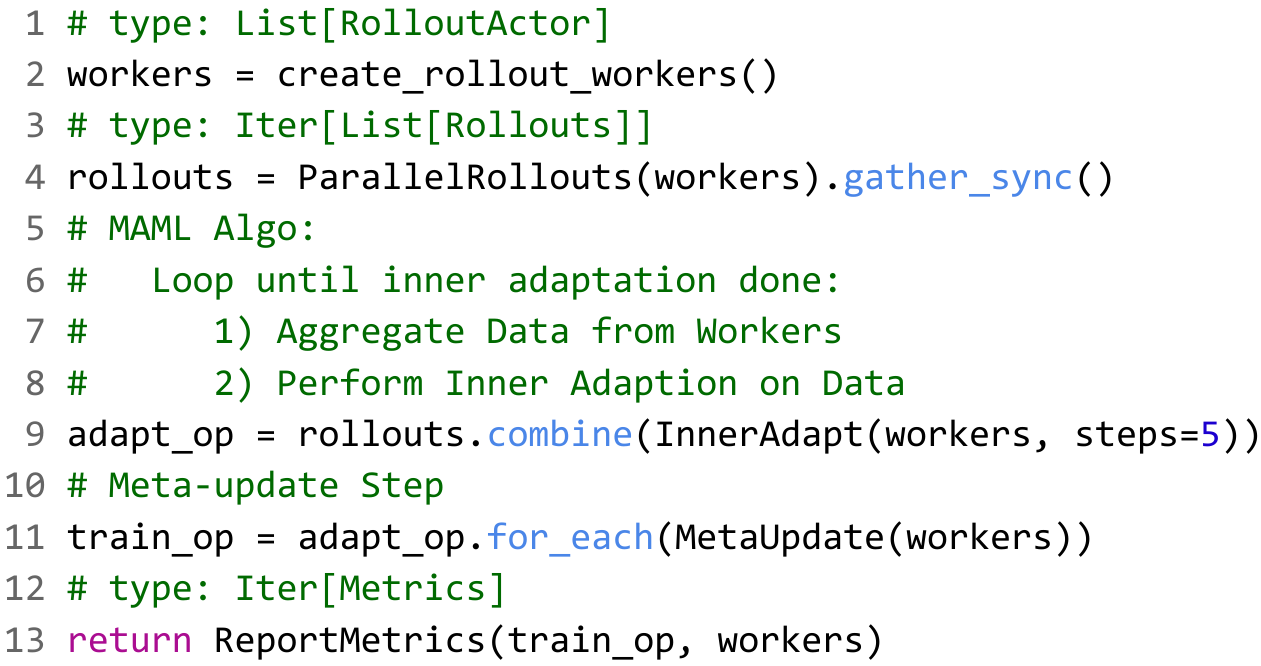}
\caption{Implementation in \flowname.}
\label{alg:maml-code}
\end{subfigure}
\caption{\revise{Dataflow and implementation of the MAML algorithm.}}
\vspace{-10pt}
\end{figure}

\code{maml-code} concisely expresses MAML's dataflow (also shown in Figure \ref{fig:maml}) \cite{finn2017modelagnostic}. The MAML dataflow involves nested optimization loops; workers collect pre-adaptation data, perform inner adaptation (i.e., individual optimization calls to an ensemble of models spread across the workers), and collect post-adaptation data. Once inner adaptation is complete, the accumulated data is batched together to compute the meta-update step, which is broadcast to all workers. 
% \zhanghao{TODO: descriptions}

\section{Comparison of Implementations in \flowname and RLlib}
\label{sec:code-comparison}
% TODO: detailed code comparison of A3C and ApeX

In this section we report the detailed code comparison of our \flowname and the original RLlib. \lst{a3c-detailed} and \lst{a3c-detailed-origin} are the detailed implementation of A3C in \flowname and RLlib, respectively. Note that the detailed implementation in~\lst{a3c-detailed} is exactly the same as we shown before in~\code{a3c-code}, but RLlib implementation is much more complicated as the intermixing of the control and data flow.
In~\lst{apex-detailed} and~\lst{apex-detailed-origin}, we also show the detailed implementation of Ape-X algorithm in our~\flowname and~\libname respectively, which also indicates the simplicity, readability and flexibility of our~\flowname.
\begin{mycode}
\caption{Detailed A3C in \flowname.}
\label{lst:a3c-detailed}
\begin{minted}{python}
# type: List[RolloutActor]
workers = create_rollout_workers()
# type: Iter[Gradients]
grads = ParallelRollouts(workers)
    .par_for_each(ComputeGradients())
    .gather_async()
# type: Iter[TrainStats]
apply_op = grads
    .for_each(ApplyGradients(workers))
# type: Iter[Metrics]
return ReportMetrics(apply_op, workers)
\end{minted}
\end{mycode}

\begin{mycode}
\caption{Detailed A3C in original RLlib.}
\label{lst:a3c-detailed-origin}
\begin{minted}{python}
# Create timers
apply_timer = TimerStat()
wait_timer = TimerStat()
dispatch_timer = TimerStat()

# Create training information
num_steps_sampled = 0

# type: List[RolloutActor]
workers = create_rollout_workers()

# Get weights from the local rollout actor
local_worker = workers.local_worker()
weights = local_worker.get_weights()

# Put weights in raylet (distributed storage)
weights = ray.put(weights)

# type: Dict[obj_id, RolloutActor]
pending_gradients = dict()

# Get the remote rollout actors
remote_worker = workers.remote_workers()

# Issue gradient computation tasks
for worker in remote_worker:
    # Set weight on remote rollout actor
    worker.set_weights.remote(weights)
    # Collect samples from the remote rollout actor
    samples = worker.sample.remote()

    # Kick off gradient computation 
    future = worker.compute_gradients.remote(samples)

    # Map the object id to rollout actor
    pending_gradients[future] = worker

# Start training loop
while pending_gradients:
    # Record the time to wait gradient
    with wait_timer:
        # Get the list of the futures
        futures = list(pending_gradients.keys())

        # Wait for one actor to complete
        wait_results = ray.wait(futures, 
                            num_returns=1)

        # Get the ready future
        ready_list = wait_results[0]
        future = ready_list[0]

        # Get and free the gradient and training infos
        # from the raylet (maybe on the remote worker)
        gradient, info = ray_get_and_free(future)

        # Pop the used gradient from the map
        worker = pending_gradients.pop(future)

    # Check the validation of the gradient
    if gradient is not None:
        # Record the time for gradient apply
        with apply_timer:
            # Apply the gradient on the local worker
            local_worker = workers.local_worker()
            local_worker.apply_gradients(gradient)

        # Record the metrics from the worker
        num_steps_sampled += info["batch_count"]

    # Record the time to set new weight on the worker 
    # and launch gradient computation task
    with dispatch_timer:
        # Get the weight on local rollout actor
        local_worker = workers.local_worker()
        weights = local_worker.get_weights()
        
        # Set weight on the rollout actor
        worker.set_weights.remote(weights)

        # Sample rollouts on the rollout actor
        samples = worker.sample.remote()
        # Launch gradient computation task on the worker
        future = worker.compute_gradients.remote(samples)

        # Map the new object id to the corresponding worker
        pending_gradients[future] = worker
\end{minted}
\end{mycode}

\begin{mycode}
\caption{Detailed Ape-X in \flowname.}
\label{lst:apex-detailed}
\begin{minted}{python}
# type: List[RolloutActor]
workers = create_rollout_workers()

# Create a number of replay buffer actors.
replay_actors = create_colocated(ReplayActor)

# Start the learner thread.
learner_thread = LearnerThread(workers.local_worker())
learner_thread.start()

# We execute the following steps concurrently:
# (1) Generate rollouts and store them in our replay buffer actors. Update
# the weights of the worker that generated the batch.
rollouts = ParallelRollouts(workers, mode="async", num_async=2)
store_op = rollouts \
    .for_each(StoreToReplayBuffer(actors=replay_actors))

# Only need to update workers if there are remote workers.
store_op = store_op.zip_with_source_actor() \
    .for_each(UpdateWorkerWeights(workers))

# (2) Read experiences from the replay buffer actors and send to the
# learner thread via its in-queue.
replay_op = Replay(actors=replay_actors, num_async=4) \
    .zip_with_source_actor() \
    .for_each(Enqueue(learner_thread.inqueue))

# (3) Get priorities back from learner thread and apply them to the
# replay buffer actors.
update_op = Dequeue(learner_thread.outqueue) \
    .for_each(UpdateReplayPriorities()) \
    .for_each(TrainOneStep(workers))

# Execute (1), (2), (3) asynchronously as fast as possible. Only output
# items from (3) since metrics aren't available before then.
merged_op = Concurrently(
    [store_op, replay_op, update_op], mode="async", output_indexes=[2])

return ReportMetrics(merged_op, workers)
\end{minted}
\end{mycode}

\begin{mycode}
\caption{Detailed Ape-X in original RLlib. We leave out some of the configurable argument for simplicity.}
\label{lst:apex-detailed-origin}
\begin{minted}{python}
# type: List[RolloutActor]
workers = create_rollout_workers()

# Create a learner thread in the main driver to handle 
# the asynchronous training
local_worker = workers.local_worker()
learner = LearnerThread(local_worker)

# Start the learner thread and wait for the input
learner.start()

# Create replay actor handling the replay buffer
# create_located: create multiple colocated replay actor
# in the same machine as main driver
replay_actors = create_colocated(ReplayActor)

# Create timers
timers = {
    k: TimerStat()
    for k in [
        "put_weights", "get_samples", "sample_processing",
        "replay_processing", "update_priorities", "train", "sample"
    ]
}

# Create training information
num_weight_syncs = 0
num_samples_dropped = 0
learning_started = False

# Number of worker steps since the last weight update
steps_since_update = dict()

# Create manager for replay
replay_tasks = TaskPool()
# Kick off replay tasks for local gradient updates
for actor in replay_actors:
    # Start replay task on remote replay actors
    for _ in range(REPLAY_QUEUE_DEPTH):
        replay_task = actor.replay.remote()
        # add replay task into the manager
        replay_tasks.add(actor, replay_task)

# Create manager for sampling
sample_tasks = TaskPool()

# Get weights of local worker
local_worker = workers.local_worker()
weights = local_worker.get_weights()

# Kick off async background sampling and set the weights 
# on remote rollout actors
remote_workers = workers.remote_workers()
for worker in remote_workers:
    # Set weights 
    worker.set_weights.remote(weights)
    # Initialize training info for the rollout actor
    steps_since_update[worker] = 0
    for _ in range(SAMPLE_QUEUE_DEPTH):
        # Start sample_with_count task on remote worker
        sample_with_count_task = worker.sample_with_count.remote()
        # Add task in to the sample task manager
        sample_tasks.add(worker, sample_with_count_task)

# Optimize the model for one step
def step(self):
    # Check the availability of the asynchronous learner thread
    # and the remote rollout actors
    assert self.learner.is_alive()
    assert len(self.workers.remote_workers()) > 0

    # Record the start time for training info
    start = time.time()

    # Create variables for training
    sample_timesteps, train_timesteps = 0, 0
    weights = None

    # Record the sampling and processing step
    with timers["sample_processing"]:
        # Check the completed sampling task in the sampling manager (TaskPool)
        completed = list(sample_tasks.completed())

        # Gather the train info, counts of samples
        counts = ray_get_and_free([c[1][1] for c in completed])
        
        # Update training information and weights
        for i, (worker, (sample_batch, count)) in enumerate(completed):
            # Update training information
            sample_timesteps += counts[i]

            # Randomly choose one replay actor and send data to it
            random_replay_actor = random.choice(replay_actors)
            random_replay_actor.add_batch.remote(sample_batch)

            # Update train info
            steps_since_update[worker] += counts[i]

            # Update weights on remote rollout worker if needed
            if steps_since_update[worker] >= MAX_WEIGHT_SYNC_DELAY:
                # Note that it's important to pull new weights once
                # updated to avoid excessive correlation between actors
                if weights is None or learner.weights_updated:
                    learner.weights_updated = False

                    # Record time for putting weights
                    with timers["put_weights"]:
                        # Put local weights in raylet
                        local_worker = workers.local_worker()
                        local_weights = local_worker.get_weights()
                        weights = ray.put(local_weights)
                
                # Set weights on the remote rollout worker 
                worker.set_weights.remote(weights)

                # Update train info
                num_weight_syncs += 1
                steps_since_update[worker] = 0

            # Kick off another sample request
            sample_with_count = worker.sample_with_count.remote()
            # Add the task into the sample manager
            sample_tasks.add(worker, sample_with_count)

    # Record the time for replay and processing
    with self.timers["replay_processing"]:
        for actor, replay in replay_tasks.completed():
            # Start another replay task for each completed one
            replay_task = actor.replay.remote()
            replay_tasks.add(actor, replay_task)
            
            # Check the input queue of the learner
            if learner.inqueue.full():
                num_samples_dropped += 1
            else:
                # Record the get sample time
                with self.timers["get_samples"]:
                    samples = ray_get_and_free(replay)

                # Defensive copy against plasma crashes
                learner.inqueue.put((actor, samples.copy()))

    # Record the time for priorities update
    with timers["update_priorities"]:
        # Get output from the leaner to update replay priorities on 
        # the remote rollout actors and training info
        while not learner.outqueue.empty():
            # Fetch output from the asynchronous learner
            output = learner.outqueue.get()
            actor, priority_dict, count = output

            # Update the priorities on the remote actors
            actor.update_priorities.remote(priority_dict)
            train_timesteps += count

    # Calculate the time information
    time_delta = time.time() - start

    # Collect metrics for training
    timers["sample"].push(time_delta)
    timers["sample"].push_units_processed(sample_timesteps)
    if train_timesteps > 0:
        learning_started = True
    if learning_started:
        timers["train"].push(time_delta)
        timers["train"].push_units_processed(train_timesteps)
    
    # Update training info
    num_steps_sampled += sample_timesteps
    num_steps_trained += train_timesteps
\end{minted}
\end{mycode}

\end{document}